\documentclass[sigconf, nonacm, pdfa]{acmart}

\newcommand\vldbdoi{XX.XX/XXX.XX}
\newcommand\vldbpages{XXX-XXX}
\newcommand\vldbvolume{17}
\newcommand\vldbissue{5}
\newcommand\vldbyear{2024}
\newcommand\vldbauthors{\authors}
\newcommand\vldbtitle{\shorttitle} 
\newcommand\vldbavailabilityurl{https://github.com/Happy2Git/FATS_supplement}
\newcommand\vldbpagestyle{empty} 
\usepackage{dsfont}
\usepackage{balance}
\usepackage{amsmath,amsfonts,amsthm}

\usepackage{amssymb}
\usepackage[ruled,vlined,linesnumbered]{algorithm2e}
\usepackage{pifont}
\newcommand{\cone}{\ding{192}}
\newcommand\ctwo{\ding{193}}
\newcommand\cthree{\ding{194}}
\newcommand\cfour{\ding{195}}
\newcommand\cfive{\ding{196}}

\usepackage{mathtools}
\usepackage{graphicx}
\theoremstyle{definition}
\newtheorem{assumption}{Assumption}
\newtheorem{theorem}{Theorem}
\newtheorem{definition}{Definition}
\newtheorem{corollary}{Corollary}
\newtheorem{lemma}{Lemma}
\newtheorem{remark}{Remark}
\newtheorem{claim}{Claim}

\usepackage{color}
\allowdisplaybreaks[4]

\DeclareMathOperator*{\TV}{\operatorname{TV}}
 
\usepackage{arydshln}
\usepackage{caption}
\usepackage{subcaption}
\usepackage{multirow}
\begin{document}
\title{Communication Efficient and Provable Federated Unlearning}

\author{Youming Tao}
\affiliation{%
  \institution{Shandong University}
  \state{P.R. China}
}
\email{ym.tao99@mail.sdu.edu.cn}
\authornote{Youming Tao and Cheng-Long Wang contributed equally to this work. Part of the work was done when Youming Tao was a research intern at KAUST}

\author{Cheng-Long Wang}
\orcid{0000-0003-2391-0923}
\affiliation{%
  \institution{KAUST}
  \country{Saudi Arabia}
}
\email{chenglong.wang@kaust.edu.sa}
\authornotemark[1]

\author{Miao Pan}
\affiliation{%
  \institution{University of Houston}
  \country{U.S.A.}
}
\email{mpan2@uh.edu}

\author{Dongxiao Yu}
\affiliation{%
  \institution{Shandong University}
  \state{P.R. China}
}
\email{dxyu@sdu.edu.cn}
\authornote{Corresponding author: Dongxiao Yu, Di Wang.}

\author{Xiuzhen Cheng}
\affiliation{%
  \institution{Shandong University}
  \state{P.R. China}
}
\email{xzcheng@sdu.edu.cn}

\author{Di Wang}
\affiliation{%
  \institution{KAUST}
  \country{Saudi Arabia}
}
\email{di.wang@kaust.edu.sa}
\authornotemark[2]

\begin{abstract}
We study federated unlearning, a novel problem to eliminate the impact of specific clients or data points on the global model learned via federated learning (FL). This problem is driven by the right to be forgotten and the privacy challenges in FL. We introduce a new framework for exact federated unlearning that meets two essential criteria: \textit{communication efficiency} and \textit{exact unlearning provability}. To our knowledge, this is the first work to tackle both aspects coherently. We start by giving a rigorous definition of \textit{exact} federated unlearning, which guarantees that the unlearned model is statistically indistinguishable from the one trained without the deleted data. We then pinpoint the key property that enables fast exact federated unlearning: total variation (TV) stability, which measures the sensitivity of the model parameters to slight changes in the dataset. Leveraging this insight, we develop a TV-stable FL algorithm called \texttt{FATS}, which modifies the classical \texttt{\underline{F}ed\underline{A}vg} algorithm for \underline{T}V \underline{S}tability and employs local SGD with periodic averaging to lower the communication round. We also design efficient unlearning algorithms for \texttt{FATS} under two settings: client-level and sample-level unlearning. We provide theoretical guarantees for our learning and unlearning algorithms, proving that they achieve exact federated unlearning with reasonable convergence rates for both the original and unlearned models. We empirically validate our framework on 6 benchmark datasets, and show its superiority over state-of-the-art methods in terms of accuracy, communication cost, computation cost, and unlearning efficacy.
\end{abstract}

\maketitle

\pagestyle{\vldbpagestyle}
\begingroup\small\noindent\raggedright\textbf{PVLDB Reference Format:}\\
\vldbauthors. \vldbtitle. PVLDB, \vldbvolume(\vldbissue): \vldbpages, \vldbyear.\\
\href{https://doi.org/\vldbdoi}{doi:\vldbdoi}
\endgroup
\begingroup
\renewcommand\thefootnote{}\footnote{\noindent
This work is licensed under the Creative Commons BY-NC-ND 4.0 International License. Visit \url{https://creativecommons.org/licenses/by-nc-nd/4.0/} to view a copy of this license. For any use beyond those covered by this license, obtain permission by emailing \href{mailto:info@vldb.org}{info@vldb.org}. Copyright is held by the owner/author(s). Publication rights licensed to the VLDB Endowment. \\
\raggedright Proceedings of the VLDB Endowment, Vol. \vldbvolume, No. \vldbissue\ %
ISSN 2150-8097. \\
\href{https://doi.org/\vldbdoi}{doi:\vldbdoi} \\
}\addtocounter{footnote}{-1}\endgroup

\ifdefempty{\vldbavailabilityurl}{}{
\vspace{.3cm}
\begingroup\small\noindent\raggedright\textbf{PVLDB Artifact Availability:}\\
The source code, data, and/or other artifacts have been made available at \url{\vldbavailabilityurl}.
\endgroup
}

\section{Introduction}

The proliferation of personal data collection and processing by various entities poses a serious threat to the public’s data privacy. To reconcile users’ data privacy and the need for data analytics in intelligent applications, Federated Learning (FL)~\cite{mcmahan2017communication} has emerged as a promising paradigm for collaborative machine learning at edge. FL allows multiple edge devices (or clients\footnote{In this paper, we use the term device and client interchangeably.}) to cooperatively learn a global model with the coordination of a central edge server. By keeping users’ data locally and only exchanging model updates, FL mitigates the risk of direct privacy leakage or disclosures. However, this is not enough to address the privacy issues as adversaries can still deduce sensitive user information from the shared models. Therefore, users require the ability to erase some of their private data from the global model, which is known as \textit{the right to be forgotten} and has been supported by several legal regulations such as GDPR~\cite{voigt2017eu} and CCPA~\cite{harding2019understanding}. This is particularly relevant in the setting of edge computing, where edge devices such as smartphones, tablets, smartwatches, or sensors generate and process massive amounts of diverse and confidential data from users. For example, some users may wish to remove their health records from a wearable device or their location history from a navigation app. However, simply deleting data from edge devices may not guarantee the right to be forgotten, as the data may have been used to train collaborative learning models that are distributed among multiple edge devices and the server. This implies that the data may still affect the model’s outcomes or actions or even be recovered by malicious actors \cite{song2020analyzing,wang2019beyond,zhu2019deep} Hence, it is essential to develop \textit{machine unlearning} \cite{bourtoule2021machine} methods to endow FL models with the capability to unlearn requested data, i.e., to eliminate the impact of some specific data from a trained FL model. Moreover, data removal from a trained model is also beneficial in other scenarios, such as countering data poisoning attacks and correcting data errors~\cite{nguyen2022survey}.

Despite its importance and necessity, machine unlearning is not a trivial task, even in the centralized setting where a single entity stores and processes the data and the model. It requires efficient algorithms that can update the model parameters without retraining from scratch, as well as verify the effectiveness and soundness of the unlearning process. Moreover, machine unlearning becomes exceedingly formidable in the federated setting. Machine unlearning in the federated setting, known as \textit{federated unlearning}, has to cope with several additional factors. 

(I) Communication efficiency is paramount for federated unlearning, as communication is the primary bottleneck of FL \cite{kairouz2021advances}. Hence, unlearning methods that incur frequent communication between the server and the clients, or model retraining from scratch, are highly impractical.

(II) The availability of training data is severely limited in federated unlearning, as the data are distributed across clients and not divulged to the server. Therefore, unlearning methods that depend on accessing or reusing the unlearned data are inapplicable in the federated setting.

(III) Besides sample-level unlearning in the central setting, federated unlearning also entails client-level unlearning, which aims to eliminate the influence of specific clients from the FL model. This is because the edge environment is unstable and clients may withdraw from the system at any time. 

(IV) Federated unlearning requires provable guarantees for both sample and client-level unlearning, as unlearning methods should be trustworthy and capable of certifying that they effectively and completely erase the influence of the unlearned samples or clients with a reasonable convergence guarantee for the unlearned model.
 
 These factors make federated unlearning a more intricate and challenging problem than centralized machine unlearning. Consequently, most of the existing machine unlearning methods are not applicable in FL. Furthermore, existing works on federated unlearning are scarce and have many limitations. For example, most of them proposed heuristic methods without any theoretical unlearning guarantees \cite{liu2021federaser,wu2022federated,wu2022knowledge}, or noise perturbation-based algorithm with only an approximate unlearning guarantee \cite{fraboni2022sequential} (see Section \ref{sec-rw} for more details). How to enable efficient data removal from FL models with provable guarantees remains largely under-explored. 

\textbf{Our Contributions.} In this paper, we take the first step to fill the gap by presenting a general framework for communication efficient and provable federated unlearning. It is noteworthy that, compared with the previous works, our goal is to achieve \textit{exact} federated unlearning. That is, when a specific data point or client is requested to be removed, the algorithmic states are adjusted exactly to what they would have been if the data or client had never been included. Our main contributions can be summarized as follows:

 (I) We provide a rigorous definition of \textit{exact} federated unlearning, which guarantees that, for each sample or client deletion request, the unlearned model is statistically indistinguishable from the one trained without the deleted sample or client from scratch. To the best of our knowledge, this is the first formal definition of exact federated unlearning. Moreover, we show that the efficiency of exact federated unlearning depends on a key property of the FL algorithm: total variation (TV) stability (at both sample and client levels), which offers guidance for FL algorithm design.
 
 (II) To achieve TV stable algorithm for FL, we develop \texttt{FATS}, which leverages local SGD with periodic averaging for communication efficiency and incorporates an elaborately designed sub-sampling rule. Specifically, at each communication round, only part of the clients are selected to run multiple local update steps using sub-sampled partial local data. We then devise efficient sample-level and client-level unlearning algorithms for \texttt{FATS}. Informally, for $M$ clients each with $N$ data samples, and a given pair of sample and client level TV-stability parameters $\rho_S ,\rho_C\in(0, 1]$, our unlearning algorithms only retrain on  $\rho_S$ and $\rho_C$ fraction of sample and client removal requests, respectively, while achieving both exact unlearning requirement and the reasonable convergence error of $O\left(\frac{1}{\sqrt{\rho_S MN}}\right)$ in terms of average-squared gradient norm over non-convex loss and non-i.i.d. data.

 (III) Finally, we evaluate our proposed framework on 6 benchmark datasets to show its effectiveness and efficiency. We find that \texttt{FATS} matches the performance of the standard FL baselines in terms of test accuracy and convergence speed, and that the unlearning algorithms can drastically reduce the computation and communication costs for both sample and client unlearning compared with state-of-the-art federated unlearning methods. Experimental results confirm that our framework achieves the best possible learning performance and restores utility faster upon receiving the unlearning request, which provides practical guarantees that fully comply with the requirement of exact federated unlearning.

\section{Related Work}\label{sec-rw}

\textbf{Machine unlearning.} The study of machine unlearning was pioneered by \cite{cao2015towards}, where exact unlearning methods were devised for statistical query learning to guarantee the equivalence of the unlearned model and the model retrained from scratch without the deleted sample. However, their methods are limited to very structured problems. A more general framework for deep learning models was proposed by \cite{bourtoule2021machine}, which introduced the Sharded, Isolated, Sliced, Aggregated (SISA) approach. Subsequently, several works also adapted the SISA framework to graph data, e.g., \cite{chen2022graph,wang2023inductive}. Nevertheless, these works all lack theoretical guarantees for their approaches and are not suitable for the federated learning setting. Another line of work relaxed the exact unlearning requirement and considered approximate unlearning starting from \cite{ginart2019making}, which was inspired by differential privacy \cite{dwork2006calibrating} and allowed the distribution of the unlearning model to be only close to that of the retaining model. Various works have explored approximate unlearning algorithms for different learning objectives, such as empirical risk minimization \cite{guo2019certified,neel2021descent,thudi2022unrolling} and population risk minimization \cite{sekhari2021remember}. Although they provided theoretical error bounds for their algorithms, the approximate unlearning objective implies that the deleted data was not fully removed and could still be leaked to an adversary, which makes it less desirable than exact unlearning. To bridge the theoretical gap for exact unlearning, \cite{ullah2021machine} introduced a notion of algorithmic stability, Total Variation (TV) stability, which is used for achieving exact unlearning. It designed TV-stable algorithms, analyzed the trade-offs between accuracy and unlearning efficiency, and established upper and lower bounds on the excess empirical and population risks of TV-stable learning algorithms over convex loss functions. However, its notion of TV stability is not directly transferable to the federated setting, and the design of TV-stable FL algorithms remains an open challenge. Moreover, its theoretical analysis and results are based on convex optimization with i.i.d. data, while we consider federated non-convex optimization with non-i.i.d. data in this work.

\textbf{Federated unlearning.} Unlearning in the federated learning paradigm has received relatively less attention than in the centralized paradigm. FedEraser \cite{liu2021federaser} reconstructs the unlearned global model approximately by leveraging the historical parameter updates of the participating clients. Although speeding up the unlearning process compared with retraining from scratch, it is still prohibitively expensive. Thus, more recent works have proposed approximate unlearning techniques based on knowledge distillation \cite{wu2022knowledge}, class-discriminative pruning \cite{wang2022federated}, projected gradient ascent \cite{halimi2022federated,wu2022federated}, or second-order AdaHessian optimizer \cite{liu2022right}. However, these works still suffer from several limitations: Firstly, these approximate unlearning techniques cannot completely eliminate the influence of data samples to be removed from the trained model, which may result in severe privacy leakage on data holders’ data samples, e.g., gradient leakage attacks \cite{song2020analyzing}. Secondly, most of them only consider client-level unlearning. Sample-level unlearning is also crucial and distinct from the centralized paradigm. As clients will not share their private data with the server, the federated unlearning process can only be executed on the clients. Thirdly, prior works are mostly heuristic. They only validate their method empirically without any theoretical guarantees, which makes their methods unreliable in real-world applications. To overcome these challenges, in this paper, we pursue a more ambitious goal of exact federated unlearning at both sample and client levels, and also provide rigorous theoretical analysis on both unlearning guarantees and convergence error bound. Moreover, we consider the communication bottleneck in the federated scenario, which is seldom addressed in prior works but is vital in practice.

\section{Preliminaries}\label{sec-pre}

\subsection{Federated Learning} \label{sec-pre-FL}
We consider a typical federated learning setup where $M$ clients cooperate to train a machine learning model under the coordination of a central server. Let $\mathcal{X}$ be the data universe and $\Theta$ be the model parameter space in general. Each client $k\in[M]$ has a local dataset $\mathcal{D}_k$ comprising $N$ data points $X_k^{(1)}, X_k^{(2)}, \ldots, X_k^{(N)}\in\mathcal{X}$ sampled from a client-specific local distribution $D_k$. The goal is to learn a machine learning model parameterized by $\theta\in\Theta$ that minimizes the global empirical risk function $F(\cdot)$ over the $M$ local datasets without disclosing the data to the central server due to privacy concerns. The goal can be formally described as the following empirical risk minimization (ERM) problem:
\begin{equation}\label{eq-erm}
    \min_{\theta\in\Theta}F(\theta)\coloneqq \frac{1}{M}\sum_{k=1}^M F_k(\theta), 
\end{equation}
where $F_k(\theta)$ is the local empirical risk function for client $k$ and is defined as $F_k(\theta)\coloneqq\frac{1}{N}\sum_{i=1}^N f(\theta; X_{k}^{(i)})$ with $f(\theta;X_{k}^{(i)})$ being the point-wise loss of model parameter $\theta$ for the data sample $X_{k}^{(i)}$.

\subsection{Federated Unlearning}
The goal of machine unlearning is to remove the influence of the requested data from a trained machine learning model. In the federated setting, we distinguish two types of \textit{exact} unlearning requirements --- sample-level unlearning and client-level unlearning. A pair of federated learning and unlearning procedure can be denoted as a tuple $(\mathcal{L},\mathcal{U})$, where $\mathcal{L}: \mathcal{X}^*\to\Theta\times\mathcal{S}$ is the learning algorithm that maps a set of training data points to a trained model $\theta\in\Theta$ together with an internal algorithmic state $s\in\mathcal{S}$, and $\mathcal{U}:\Theta\times\mathcal{S}\times \mathcal{X}^*\to\Theta\times\mathcal{S}$ is the unlearning algorithm that updates the learned model $\theta\in\Theta$ and internal state $s\in\mathcal{S}$ to the unlearned model $\theta^u$ and corresponding state $s^u$, given an unlearning request (either a target data point in some client or the entire dataset of a target client\footnote{In this paper, we formally refer to the data sample and client that need to be forgotten as the target sample and the target client respectively.}). We collectively refer to the output model $\theta$ and internal state $s$ as the \textit{algorithmic state}. In FedAvg algorithm with local mini-batch SGD and client sub-sampling for example, the output model is the final global model whereas the rest of intermediate local and global models, local mini-batches, subsets of clients are the internal state. We use $\mathcal{C}$ to denote the set of clients and $\mathcal{D}(\mathcal{C})$ to denote all the data points stored by the clients in $\mathcal{C}$. We may omit the client set indicator $\mathcal{C}$ and simply use $\mathcal{D}$ to denote the global dataset when it is clear from the context. For any two sets $\mathcal{Z}$ and $\mathcal{Z}^\prime$, we use $\Delta(\mathcal{Z},\mathcal{Z}^\prime)$ to denote the symmetric difference between them, {\em i.e.,} $\Delta(\mathcal{Z}, \mathcal{Z}^\prime)\coloneqq(\mathcal{Z}\setminus \mathcal{Z}^\prime)\cup(\mathcal{Z}^\prime \setminus \mathcal{Z})$. For any set $\mathcal{Z}$, we use $|\mathcal{Z}|$ to denote its cardinality. Now, we define the notions of both sample-level and client-level exact federated unlearning.
\begin{definition}[Sample-level Exact Federated Unlearning]\label{def-data-unlearning}
    A pair of federated learning and unlearning algorithms $(\mathcal{L},\mathcal{U})$ is said to satisfy sample-level exact federated unlearning if for any $\mathcal{D}(\mathcal{C}), \mathcal{D}(\mathcal{C})^\prime$ such that $\mathcal{D}(\mathcal{C})\supset\mathcal{D}(\mathcal{C})^\prime$ and $|\Delta(\mathcal{D}(\mathcal{C}),\mathcal{D}(\mathcal{C})^\prime)|=1$, and any measurable event $\mathcal{E}\subseteq\Theta\times\mathcal{S}$, it holds that  
    \begin{equation*}
        \mathbb{P}(\mathcal{L}(\mathcal{D}(\mathcal{C})^\prime)\in\mathcal{E})=\mathbb{P}(\mathcal{U}(\mathcal{L}(\mathcal{D}(\mathcal{C})), \Delta(\mathcal{D}(\mathcal{C}), \mathcal{D}(\mathcal{C})^\prime))\in\mathcal{E}).
    \end{equation*}
\end{definition}

\begin{definition}[Client-level Exact Federated Unlearning]\label{def-client-unlearning}
    A pair of federated learning and unlearning algorithms $(\mathcal{L},\mathcal{U})$ is said to satisfy client-level exact federated unlearning if for any $\mathcal{C}, \mathcal{C}^\prime$ such that $\mathcal{C}\supset\mathcal{C}^\prime$ and $|\Delta(\mathcal{C},\mathcal{C}^\prime)|=1$, and any measurable event $\mathcal{E}\subseteq\Theta\times\mathcal{S}$, it holds that 
    \begin{equation*}
        \mathbb{P}(\mathcal{L}(\mathcal{D}(\mathcal{C}^\prime))\in\mathcal{E})=\mathbb{P}(\mathcal{U}(\mathcal{L}(\mathcal{D}(\mathcal{C})), \Delta(\mathcal{D}(\mathcal{C}),\mathcal{D}(\mathcal{C}^\prime)))\in\mathcal{E}).
    \end{equation*}
\end{definition}

The above definitions imply that, for each target sample or target client to delete, the unlearned model is equivalent to the one that would have been obtained if trained on the updated setting (i.e., without the target sample or target client) from scratch, since they have exactly identical output distributions. This is the reason why we refer to this type of unlearning requirement as \textit{exact} unlearning. Furthermore, although the above definitions are for a single edit request, they can be extended for a sequence of $w$ edit requests by requiring the condition to hold recursively for every request in the sequence.

\subsection{Federated Unlearning via TV-Stability}\label{sec:3.3}
We adopt the total variation (TV) distance as a measure of discrepancy between two distributions $P$ and $Q$:
\begin{equation}
    \TV(P,Q)\coloneqq\sup_{\text{measurable set } \mathcal{Z}}|P(\mathcal{Z})-Q(\mathcal{Z})|=\frac{1}{2}\|\phi_P-\phi_Q\|_1,
\end{equation}
where the second equality is valid when distributions $P$ and $Q$ admit probability densities (denoted as $\phi_P$ and $\phi_Q$ respectively) with respect to a base measure.

Let $\mathcal{D}$ and $\mathcal{D}^\prime$ be two datasets such that $\mathcal{D}^\prime\subseteq\mathcal{D}$ is obtained by removing a data point or a client from $\mathcal{D}$. Let $P=\mathcal{L}(\mathcal{D})$ and $Q=\mathcal{L}(\mathcal{D}^\prime)$ for some randomized FL algorithm $\mathcal{L}$. To satisfy the requirement of exact federated unlearning, we need to move from $P$ to $Q$, which is an optimal transport problem. 

In a general optimal transport problem, we are given two probability distributions $P$ and $Q$ over a measurable space $\Omega$ and a cost function $c:\Omega\times\Omega\to\mathbb{R}$. The goal is to transport from $P$ to $Q$ using the minimum cost. Formally, let $\Pi(P,Q)$ denote the set of couplings (or transport plans) of $P$ and $Q$, we seek a transport plan $\pi$ which minimizes the expected cost: $\min_{\pi\in\Pi(P,Q)}\mathbb{E}_{(x,y)\sim\pi}c(x,y)$. 

In the context of federated unlearning, we aim to find a transport from $P=\mathcal{L}(\mathcal{D})$ to $Q=\mathcal{L}(\mathcal{D}^\prime)$ over the set of all possible algorithmic states $\Omega=\Theta\times\mathcal{S}$. Notably, we require the probability distribution of the entire algorithmic state, not just the output, to be exactly identical after unlearning. A trivial transport plan is re-computation that generates independent samples from $P$ and $Q$, but this is extremely inefficient as it requires a full re-training from scratch for every unlearning request. To improve the efficiency of unlearning, we should exploit the correlation between $P$ and $Q$ so that we can reuse the randomness (computation) involved in generating P when transforming $P$ to $Q$, hoping that no re-training or only a partial re-training from the middle of the learning process is sufficient for every unlearning request. For this purpose, we define the cost function for any two distinct algorithmic states, which captures the expected costs for any transport rule from $P$ to $Q$. A natural choice is to set the cost function as $c(x,y)=\begin{cases} 1, \text{if $x\neq y$}\\ 0, \text{otherwise}\end{cases}$, which means we incur one unit of computation if the algorithmic state produced by $P$ differs from the one produced by $Q$, which corresponds to a re-computation. Under this computation model, the optimal expected computation cost is $\inf_{\pi\in\Pi(P,Q)}\mathbb{E}_{(x,y)\sim\pi}[\mathds{1}\{x\neq y\}]$, which coincides with the total variation (TV) distance between $P$ and $Q$. Therefore, if we want to transport $P$ to $Q$ using the minimum computation cost, the expected computation cost is determined by the TV distance between $P$ and $Q$. \textbf{This implies that at least $1-\TV(P,Q)$ fraction of samples are representative for both $P$ and $Q$, and thus at most $O(\TV(P,Q))$ fraction of unlearning requests need to be handled by re-training}. Moreover, due to the sequential nature of the unlearning problem, when we produce $P$, i.e., the output and internal state on dataset $\mathcal{D}$, we are unaware of what $Q$ will be, since we don't know the upcoming unlearning request. \emph{Therefore, a desirable property to enable the unlearning ability for an FL algorithm is that its entire state should be close in TV distance uniformly over all possible $Q$'s.}

Motivated by the above discussion, we introduce the notion of sample-level and client-level TV-stability. This is a type of algorithmic stability that captures the closeness of the algorithmic states under different datasets. We will use this notion to guide the design and analysis of our federated learning and unlearning algorithms.
\begin{definition}[$\rho_S$-sample-level TV-stability]\label{def-sample-tv}
    An FL algorithm $\mathcal{L}$ is $\rho_S$-sample-level-TV-stable if
    \begin{equation}
        \sup_{\mathcal{D},\mathcal{D}^\prime: |\Delta(\mathcal{D},\mathcal{D}^\prime)|=1}\TV(\mathcal{L}(\mathcal{D}),\mathcal{L}(\mathcal{D}^\prime))\le\rho_S.
    \end{equation}
\end{definition}

\begin{definition}[$\rho_C$-client-level TV-stability]\label{def-client-tv}
    An FL algorithm $\mathcal{L}$ is $\rho_C$-client-level-TV-stable if
    \begin{equation}
        \sup_{\mathcal{C},\mathcal{C}^\prime: |\Delta(\mathcal{C},\mathcal{C}^\prime)|=1}\TV(\mathcal{L}(\mathcal{D}(\mathcal{C})),\mathcal{L}(\mathcal{D}(\mathcal{C^\prime})))\le\rho_C.
    \end{equation}
\end{definition}

\begin{remark}\label{rem-TV}
    For any two datasets $\mathcal{D}, \mathcal{D}^\prime$ such that $\mathcal{D}\supset\mathcal{D}^\prime$ and $|\Delta(\mathcal{D},\mathcal{D}^\prime)|=w$, if $\mathcal{L}$ is $\rho_S$-sample-level TV-stable, then by the triangle inequality of $\TV$ and repeated applications of the Definition~\ref{def-sample-tv}, we have that $\TV(\mathcal{L}(\mathcal{D}), \mathcal{L}(\mathcal{D}^\prime))\le w\cdot\rho_S$. Similarly, we have $\TV(\mathcal{L}(\mathcal{D}(\mathcal{C})), \mathcal{L}(\mathcal{D}(\mathcal{C}^\prime))\le w\cdot\rho_C$ for any $\mathcal{C}, \mathcal{C}^\prime$ such that $\mathcal{C}\supset\mathcal{C}^\prime$ and $|\Delta(\mathcal{C},\mathcal{C}^\prime)|=w$ if $\mathcal{L}$ is $\rho_C$-client-level TV-stable. 
\end{remark}

We have established that TV-distance is a sufficient stability measure for exact federated unlearning based on our theoretical analysis. However, we do not claim that it is a necessary condition, nor that it is the optimal choice among other possible stability measures. This is a highly non-trivial problem that deserves further investigation in future work.

\section{Our Algorithms}\label{sec-algo}

We propose our algorithms for federated learning and unlearning in this section. We devise a general framework that consists of a learning algorithm \texttt{FATS} and two unlearning algorithms to enable efficient data sample and client removal for \texttt{FATS}, respectively. In our algorithms, the server and clients will employ functions $\texttt{save}(\cdot)$ and $\texttt{load}(\cdot)$, which indicate saving and loading the variables to and from its local memory respectively.

\subsection{TV-Stable FL Algorithm: \texttt{FATS}}

\begin{algorithm}[tbp]
    \caption{Federated Averaging with TV-Stability:\\ \texttt{FATS}($t_0$, $T$, $E$, $\eta$, $\rho_S$, $\rho_C$)}
    \label{Algo-FedLearn}
    \textbf{Input:} Start iteration $t_0$, time horizon $T$, local iteration number $E$, learning rate $\eta$, TV-stability parameters $\rho_S$, $\rho_C$.
    
    $K\gets\frac{\rho_C\cdot E\cdot M}{T}, b\gets\frac{\rho_S\cdot N}{\rho_C\cdot E}$ \;
    \If {$t_0 \notin \mathcal{I}_E$}
    {
        The server performs \texttt{Load}($\mathcal{P}^{(t_0)}$) \;
        The server informs each client $k\in \mathcal{P}^{(t_0)}$ to perform \texttt{Load}($\theta_k^{(t_0)}$) \;
    }
    \For {$t\gets t_0,\ldots,T$}
    {
            \If {$t\in\mathcal{I}_E$}
            {
                The server samples a multiset of clients $\mathcal{P}^{(t)}$ of size $K$ with replacement \;
                The server performs \texttt{save}($\mathcal{P}^{(t)}$), \texttt{load}($\theta^{(t-1)}$) \;
                The server broadcasts index $t$ and the latest model $\theta^{(t-1)}$ to all clients in $\mathcal{P}^{(t)}$ such that $\theta_k^{(t-1)}=\theta^{(t-1)}, \forall k\in \mathcal{P}^{(t)}$ \;
            }
            \For {each client $k\in \mathcal{P}^{(t)}$ {\rm (in parallel)}}
            {
                Sample a mini-batch $\mathcal{B}_{k}^{(t)}$ of size $b$ uniformly at random without replacement \;
                $\tilde{g}_k^{(t)}\gets\frac{1}{b}\sum_{X\in \mathcal{B}_{k}^{(t)}}\nabla f(\theta_k^{(t-1)};X)$ \;
                $\theta_k^{(t)}\gets\theta_k^{(t-1)}-\eta\cdot \tilde{g}_k^{(t)}$ \;
                Perform \texttt{Save}($t$, $\mathcal{B}_k^{(t)}$, $\theta_k^{(t)}$) \;
            }
            \If {$t \operatorname{mod} E = 0$}
            {
                Each client $k\in \mathcal{P}^{(t)}$ sends its latest local model $\theta_k^{(t)}$ to the server \;
                The server aggregates the received local models as $\theta^{(t)}\gets\frac{1}{K}\sum_{k\in \mathcal{P}^{(t)}}\theta_{k}^{(t)}$ \;
                The server performs \texttt{save}($\theta^{(t)}$) \;
            }
    }
    \textbf{return} global model $\theta^{(T)}$ \;
\end{algorithm}

Based on our previous discussions in Section \ref{sec:3.3}, 
it is sufficient to design a TV-stable learning algorithm. Specifically, our objective is to achieve both $\rho_S$-sample-level TV-stability and $\rho_C$-client-level TV-stability simultaneously for any given stability parameters $\rho_S$ and $\rho_C$. For this purpose, we propose a federated learning algorithm called \texttt{FATS}, which extends the classical \texttt{FedAvg} algorithm~\cite{mcmahan2017communication} for TV stability. Due to the widespread usage of \texttt{FedAvg}, \texttt{FATS} can be easily integrated into existing systems. The main idea for \texttt{FATS} to achieve sample-level and client-level TV stability simultaneously is to elaborately adjust the number of clients sampled per round and the mini-batch size per iteration. We describe \texttt{FATS} in Algorithm~\ref{Algo-FedLearn}.  \texttt{FATS} operates in a synchronized manner, dividing the learning process into $T$  time steps $t=1, 2, \ldots, T$. These $T$ time steps are further grouped into $R$ communication rounds $r=1, 2, \ldots, R$, each consisting of $E$ iterations of local updates, such that $T=R\cdot E$. The $r$-th communication round covers iterations from $(r-1)\cdot E+1$ to $r\cdot E$. We denote by $\mathcal{I}_E=\{sE+1|s=0, 1,2,\ldots, \lfloor\frac{T-1}{E}\rfloor\}$ the set of time steps that mark the start of a communication round. In each communication round, \texttt{FATS} performs the following three steps. 

\paragraph{STEP 1.} At the beginning of each communication round, the server randomly draws a multiset of $K$ clients with replacement and broadcasts the latest global model to them (steps 7-10). It is possible for a client to be activated multiple times in a single communication round. Let $\mathcal{C}^{(t)}$ denote the chosen client multiset that allows repetitions. Note that $\mathcal{C}^{(t)}$ is only defined for each $t\in\mathcal{I}_E$. For convenience, we use $\mathcal{P}^{(t)}$ for every $t\in[T]$ in the algorithm to denote the most recent selected clients at time step $t$, i.e., $\mathcal{P}^{(t)}\coloneqq\mathcal{C}^{(n)}$, where $n=\max\{t^\prime|t^\prime\le t, t^\prime\in\mathcal{I}_E\}$. 

\paragraph{STEP 2.} Upon receiving the global model, the selected clients update the model parameters by using mini-batch stochastic gradient descent for $E$ iterations over their local datasets. Specifically, in each iteration $t$, the selected client $k\in\mathcal{P}^{(t)}$ first samples a mini-batch $\mathcal{B}_k^{(t)}$ of size $b$ from its local dataset $\mathcal{D}_k$ without replacement and then updates the local model $\theta_k^{(t-1)}$ using gradient descent based on the mini-batch gradient $\tilde{g}_k^{(t)}$ (step 12-17). 

\paragraph{STEP 3.} After completing $E$ local updates, the selected clients upload the local model parameters to the server, who then aggregates a new global model by averaging these local models (steps 18-22). 

\begin{algorithm}[tbp]
    \caption{Sample-level Unlearning for \texttt{FATS}:\\
    \texttt{FATS-SU}($t_u$, $X_u$, $k_u$)}
    \label{Algo-FedUnlearn-D}
    \textbf{Input:} Unlearning time step $t_u$, target sample $X_u\in\mathcal{D}_{k_u}$.

    \For {$t\gets1,\ldots, t_u$}
    {
        \If {$k_u\in \mathcal{P}^{(t)}$}
        {
            perform \texttt{Load}($\mathcal{B}_{k_u}^{(t)}$) \;
            \If {$X_u\in \mathcal{B}_{k_u}^{(t)}$}
            {
                \texttt{FATS}($t_S, T, E, \eta, \rho_S, \rho_C$) \;
                \textbf{halt} \;
            }
        }
    }
\end{algorithm}

\begin{algorithm}[tbp]
    \caption{Client-level Unlearning for \texttt{FATS}:\\
    \texttt{FATS-CU}($t_u$, $k_u$)}
    \label{Algo-FedUnlearn-C}
    \textbf{Input:} Unlearning time step $t_u$, Target $k_u$ to unlearn.

    $r_u\gets \lfloor (t_u-1)/E \rfloor +1$ \;
    \For {$r\gets 1,2,\ldots,r_u$}
    {
        $t_r\gets (r-1)\cdot E+1$ \;
        \If {$k_u\in \mathcal{P}^{(t_r)}$}
        {
            \texttt{FATS}($t_C, T, E, \eta, \rho_S, \rho_C$) \;
            \textbf{halt} \;
        }
    }
\end{algorithm}

\subsection{Unlearning Algorithms for \texttt{FATS}}

We recall that \texttt{FATS} involves two sources of randomness: the server’s client sampling and the client’s data sample sampling. When a target data sample or a target client is removed, the number of available data samples for a client or the number of available clients will change accordingly. This implies that the sampling probability distribution may differ if the algorithm is run on the updated setting (without the target data sample or client). Therefore, to achieve exact unlearning, we need to first verify whether such a discrepancy occurs, which is referred to as verification. If yes, we need to rectify this discrepancy by adjusting the sampling probability measure appropriately, which is done by re-computation. The sample-level and client-level unlearning algorithms are given in Algorithm~\ref{Algo-FedUnlearn-D} and Algorithm~\ref{Algo-FedUnlearn-C} respectively.

Suppose an unlearning request (for unlearning either a sample $X_u$ of $k_u$ or a client $k_u$) is made at time step $t_u$ in round $r_u$. For verification, For verification, we need to inspect whether the sampling probability distribution changes after deleting the target sample or client. At a high-level, we verify if the current model learned from the original dataset is \textit{equally} probable to occur after the deletion for each iteration. We first consider the sample-level unlearning case. To unlearn the target sample $X_u\in\mathcal{D}_{k_u}$, we iteratively check for each time step $1\le t\le t_u$ whether $\mathcal{B}_{k_u}^{(t))}$ can be generated by learning on the updated dataset \textbf{with the same probability}. Our specific approach is that, we check whether the target sample $X_u$ has ever been used for training the current model up till $t_u$. If not, then we affirm that no discrepancy exists between $P$ and $Q$ till $t_u$, and thus no re-computation is required to perform. Otherwise, if at some iteration $t_S\le t_u$, we find the target sample was involved in the training, then the discrepancy occurs and a re-computation should then be initiated to retrain the model starting from $t_S$ (step $6$ in Algorithm~\ref{Algo-FedUnlearn-D}). The client-level unlearning case is simpler since we only need to account for the client sampling process. Specifically, we iteratively check for each round $r\le r_u$ whether the target client $k_u$ was involved. If for some round with its first iteration being $t_C$, we find $k_u$ participated in the model training, then a re-computation should then be initiated (step $6$ in Algorithm~\ref{Algo-FedUnlearn-C}). 

\section{Main Results}\label{sec-results}

\subsection{Unlearning Guarantees} 

In this part, we establish that our framework achieves both sample and client level exact unlearning. We first show that \texttt{FATS} is TV-stable. Then we show that our unlearning algorithms are valid transports that preserve the output distribution of the model parameters. Thanks to the TV-stability, the probability of re-computation is also small, which implies that our unlearning algorithms are efficient in computation and communication.

{\noindent\bf Notations:}
We introduce some notations that we will use for our unlearning analysis. For simplicity, we slightly deviate from the notations used in the algorithm description, which are elaborated in the following. For sample-level unlearning, we suppose that client $k_u$ requests to delete the target data sample $X_u$ from its local dataset $\mathcal{D}_{k_u}$ and $\mathcal{D}_{k_u}$ becomes $\mathcal{D}^\prime_{k_u}$ after the deletion. For client-level unlearning, client $k_u$ is the target client. Let $\mathcal{D}$ and $\mathcal{D}^\prime$ be the global dataset of all data points across the clients before and after the deletion, respectively. That is, $\mathcal{D}$ and $\mathcal{D}^\prime$ differ by either the target sample $X_u\in\mathcal{D}_{k_u}$ for sample unlearning, or the whole local dataset $\mathcal{D}_{k_u}$ of the target client $k_u$ for client unlearning. Let $\theta_{\mathcal{D}}$ and $\theta_{\mathcal{D}^\prime}$ be the model learnt from $\mathcal{D}$ and $\mathcal{D}^\prime$, respectively. In each communication round $r$, there are two kinds of sampling: client sampling by the server and local mini-batch sampling by each client. We use $\mathcal{P}^{[r]}$ to denote the multiset of $K$ clients that server samples in round $r$. Let $\mathcal{B}_k^{[r]}$ denote the set of all mini-batches that client $k\in[M]$ samples during the $E$ iterations in round $r$, and let $\mathcal{B}_k^{[r,i]}$ denote the specific mini-batch that client $k$ samples in the $i$-th iteration of round $r$. Then we have $\mathcal{B}_k^{[r]}=\{\mathcal{B}_k^{[r,i]}\}_{i\in[E]}$ if $k\in\mathcal{P}^{[r]}$, and $\mathcal{B}_k^{[r]}=\emptyset$ otherwise. Finally, let $\mathcal{B}^{[r]}\coloneqq\{\mathcal{B}_k^{[r]}\}_{k\in[M]}$ denote the set of all mini-batches sampled by all clients in round $r$.

In our analysis, we will use the concept of push-forward measure. Given a measurable function $f:\mathcal{X}\to \mathcal{Y}$ and a measure $\mu$ on $\mathcal{X}$, we denote by $f\#\mu$ the push-forward measure on $\mathcal{Y}$, defined as $(f\#\mu)(\mathcal{Z})=\mu(f^{-1}(\mathcal{Z}))$ for any measurable set $\mathcal{Z}\subseteq \mathcal{Y}$.

To demonstrate that our framework ensures exact unlearning, we need the following crucial lemma, which states that our federated learning algorithm \texttt{FATS} is $\TV$-stable with respect to both sample and client level.

\begin{lemma}\label{le-stability}
    For any given $\rho_S, \rho_C\in(0,1]$, \texttt{FATS} is $\min\{\rho_S, 1\}$ sample-level TV-stable and $\min\{\rho_C, 1\}$ client-level TV-stable.
\end{lemma}

The proof of Lemma~\ref{le-stability} is in Appendix~\ref{sec:le-stability}. With Lemma~\ref{le-stability}, we are ready to demonstrate that our framework ensures exact unlearning. Furthermore, we show that the probability of re-computation is bounded linearly with the number of unlearning requests and the level of stability. The unlearning guarantee is formally stated in Theorem~\ref{th-unlearning}, whose proof is in Appendix~\ref{sec:th-unlearning}.

\begin{theorem}\label{th-unlearning}
    (\texttt{FATS}, \texttt{FATS-SU}) satisfies sample-level exact federated unlearning. (\texttt{FATS}, \texttt{FATS-CU}) satisfies client-level exact federated unlearning. Moreover, the probability of re-computation for $w$ number of sample-level or client-level unlearning requests is at most $\rho_S \cdot w$ or $\rho_C\cdot w$, respectively.
\end{theorem}

\subsection{Convergence Analysis}\label{sec:convergence}
In this part, we provide convergence guarantees for our proposed FL algorithm \texttt{FATS}. We show that the global model converges to an approximate stationary point of the empirical risk function under mild assumptions. We also show that the unlearned models preserve the accuracy of the original models, while achieving exact unlearning. These results demonstrate the effectiveness of our federated unlearning framework.

{\noindent\bf Notations:} Recall that, in Algorithm~\ref{Algo-FedLearn}, the server first randomly selects a multiset $\mathcal{P}^{(t)}$ of clients at the beginning of each round and then only the selected clients perform local updates. This introduces some technical challenges in the analysis since $\mathcal{P}^{(t)}$ varies every $E$ time steps. To overcome this difficulty, we consider analyzing an alternative equivalent procedure where the server always activates all clients at the beginning of each round and then only aggregates the updated parameters from the clients in multiset $\mathcal{P}^{(t)}$ to generate the latest global model. Specifically, the updating scheme of the alternative procedure can be described as: for all $k\in[M]$ and any time step $t\in[T]$, 
\begin{flalign}
    &\nu^{(t)}_k=\theta^{(t-1)}_k-\eta\tilde{g}_k^{(t)},\\
    &\theta^{(t)}_k=
    \begin{cases}
    \nu^{(t)}_k, &\text{if $t \operatorname{mod} E \neq 0$},\\
    \frac{1}{K}\sum_{k\in\mathcal{P}^{(t)}}\nu_k^{(t)}, &\text{if $t \operatorname{mod} E = 0$}.
    \end{cases}
\end{flalign} 
For ease of exposition, we introduce some notations that will facilitate the analysis. Recall that, in Algorithm~\ref{Algo-FedLearn}, we use $\tilde{g}_k^{(t)}$ to denote the stochastic mini-batch gradient of client $k$ in iteration $t$. We will use $g_k^{(t)}$ to denote the full gradient calculated from the entire local dataset at client $k$ in iteration $t$, i.e., $g_k^{(t)}\coloneqq\nabla F_k(\theta_k^{(t)})$. Moreover, in Algorithm~\ref{Algo-FedLearn}, $\theta^{(t)}$ is the global model that is aggregated among the selected clients in $\mathcal{P}^{(t)}$ every $E$ iterations at the end of each communication round. Thus, $\theta^{(t)}$ there is only defined for $t$ that is a multiple of $E$. In what follows, we extend the definition of $\theta^{(t)}$ for $\forall$ $t\in[T]\cup\{0\}$ as $\theta^{(t)}\coloneqq\frac{1}{K}\sum_{k\in\mathcal{P}^{(t)}}\theta_k^{(t)}$, which we also call the virtual average model.

Throughout our convergence analysis, we will use the following standard assumptions for loss function $f$.

\begin{assumption}[$L$-smoothness]\label{asp-smoothness}
 The loss function $f(\cdot; X)$ is $L$-smooth, \emph{i.e.}, for any given data sample $X$ and $\forall \theta_1,\theta_2\in\mathbb{R}^d$, we have $\|\nabla f(\theta_1;X)-\nabla f(\theta_2;X)\|_2\le L\|\theta_1-\theta_2\|$.  
\end{assumption}

\begin{assumption}[Bounded Local Variance]\label{asp-localvar}
    For each local dataset $\mathcal{D}_k$, $k\in[M]$, we can sample an independent mini-batch $\mathcal{B}_k$ with $|\mathcal{B}_k|=b$ and compute an unbiased stochastic gradient $\tilde{g}_k$ defined as $\frac{1}{b}\sum_{X\in \mathcal{B}_{k}}\nabla f(\theta;X)$ for $\forall \theta$, with its variance bounded from above as $\mathbb{E}_{\mathcal{B}_k}[\|\tilde{g}_k-g_k\|_2^2]\le\frac{G^2}{b}$,
    where $g_k=\mathbb{E}[\tilde{g}_k]=\frac{1}{N}\sum_{X\in\mathcal{D}_k}\nabla f(\theta;X)$.
\end{assumption}

We introduce the following notion of gradient diversity to quantify the dissimilarity between gradients of local empirical risk functions during the learning process.

\begin{definition}[Gradient Dissimilarity]
    We define the following quantity as gradient diversity among all the clients at the $t$-th learning iteration: $\Lambda(\theta^{(t)})\coloneqq\frac{\frac{1}{M}\sum_{k=1}^M\|\nabla F_k(\theta^{(t)}_k)\|_2^2}{\|\frac{1}{M} \sum_{k=1}^M \nabla F_k(\theta^{(t)}_k)\|_2^2}$.
\end{definition}

Clearly, for any $t$, $\Lambda(\theta^{(t)})\ge 1$. And $\Lambda(\theta^{(t)})=1$ if and only if all the local gradients $\nabla F_k(\theta_k^{(t)})$'s are equal. In the following assumption, we introduce the quantity $\lambda$, which is the upper bound on the gradient diversity and measures the degree of heterogeneity. 

\begin{assumption}[Bounded heterogeneity]\label{asp-heterogeneity}
    There is a common upper bound $\lambda$ on the gradient diversity among local empirical risk functions, such that for any $t\in[T]$, we have $\Lambda(\theta^{(t)})\le\lambda$.
\end{assumption}

With these assumptions, in the following Lemma~\ref{le-convergence}, we bound the average-squared gradient norm for \texttt{FATS} in general. The proof of Lemma~\ref{le-convergence} can be found in Appendix~\ref{sec:le-convergence}.

\begin{lemma}\label{le-convergence} 
    Under Assumption~\ref{asp-smoothness},~\ref{asp-localvar} and~\ref{asp-heterogeneity}, if the learning rate $\eta$  is small enough and satisfies the following condition:
    \begin{equation}\label{eq-rate_condition}
        -\frac{\eta}{2}+\eta^3L^2\lambda E(E-1)+\frac{\eta^2\lambda L}{2}<0.
    \end{equation}
    Then the average-squared gradient norm is bounded as
    \begin{flalign}
        &\frac{1}{T}\sum_{t=1}^{T}\mathbb{E}[\|\nabla F(\theta^{(t-1)})\|_2^2]\notag\\
        &\le\frac{2(F(\theta^{(0)})-F^*)}{\eta T}+\frac{\eta^2L^2G^2E(K+1)}{Kb}+\frac{\eta LG^2}{Kb}\label{eq-loss-bound0}\\
        &=\frac{2(F(\theta^{(0)})-F^*)}{\eta T}+\frac{\eta^2 L^2G^2E(\rho_CEM+T)}{\rho_SMN}+\frac{\eta LG^2T}{\rho_{S}MN},\label{eq-loss-bound}
    \end{flalign}
    where $F^*$ is the global minimum value. 
\end{lemma}

In Lemma~\ref{le-convergence}, we did not specify the choice of learning rate $\eta$ and per-round local iteration number $E$, and the obtained bound does not imply any convergence result for \texttt{FATS}. Now, we discuss the choice of $\eta$ and $E$ for ensuring convergence. To ensure convergence, one can choose $\eta=O\left(\frac{1}{T}\right)$. In this case, the condition~\eqref{eq-rate_condition} reduces to $\frac{E}{T}\le O\left(\sqrt{\frac{1}{\lambda}}\right)$. That is, the number of local iterations should be dependent on the data heterogeneity and generally decreases as the data heterogeneity becomes larger. Such a constraint on $E$ is quite intuitive: when the data are heterogeneous, the average of the minimizers of $F_1, \ldots, F_K$ can be very different from the minimizer of $F$. If $E$ is set too large, then each $\theta^{(t)}_k$ can converge to the minimizer of $F_k$, which makes the algorithm diverge. 

In the next Theorem~\ref{th-convergence}, we formalize the above discussion and provide the convergence rate for \texttt{FATS}. The proof of Theorem~\ref{th-convergence} is in Appendix~\ref{sec:th-convergence}.

\begin{theorem}\label{th-convergence}
    Define $\Gamma\coloneqq\frac{G^2}{L(F(\theta^{(0)}-F^*)\rho_SMN}$. Under Assumption~\ref{asp-smoothness},~\ref{asp-localvar} and~\ref{asp-heterogeneity}, if we choose $\eta=\frac{1}{L\sqrt{\Gamma}T}$, and let $T\ge\frac{2\lambda}{\sqrt{\Gamma}}$, then condition~\eqref{eq-rate_condition} reduces to $\frac{E(E-1)}{T^2}<\frac{\Gamma}{4\lambda}$. Therefore, by requiring  $\frac{E}{T}<\frac{1}{2}\sqrt{\frac{\Gamma}{\lambda}}$, we obtain the average-squared  gradient norm bound of 
    \begin{flalign}\label{eq-general-convergence}
        &\frac{1}{T}\sum_{t=1}^{T}\mathbb{E}[\|\nabla F(\theta^{(t-1)})\|_2^2]\notag\\
        &\le \frac{3\sqrt{LG^2(F(\theta^{(0)})-F^*)}}{\sqrt{\rho_S MN}}+L(F(\theta^{(0)})-F^*)\frac{E}{T}(\frac{\rho_C ME}{T}+1).
    \end{flalign}
\end{theorem}

\begin{remark}
    The above bound consists of two terms, for which we have some observations in order:
    
    \noindent (I) The first term can be seen as the cost for achieving stability, which scales as $O\left(\frac{1}{\sqrt{\rho_S MN}}\right)$ and thus is a non-vanishing term. If the per-client sample number $N$ and the client number $M$ are large enough, the cost can be made as small as possible. Though non-vanishing, we claim that this kind of stability cost is reasonable and it also appears in the existing centralized unlearning results, e.g., \cite{ullah2021machine}. In fact, the algorithmic TV-stability means that the performance of an FL algorithm on similar data sets is somewhat indistinguishable. If the accuracy of an FL algorithm is very high (i.e., the average-squared gradient norm converges to 0), then this kind of indistinguishability will be broken. Therefore, in order to achieve algorithmic stability, it is acceptable and necessary to sacrifice a certain accuracy. 
    
    \noindent (II) It is worth noting that our stability cost only depends on the sample-level stability $\rho_S$ and has nothing to do with client-level stability $\rho_C$. This is because our stability cost comes from balancing the first and last terms in the \eqref{eq-loss-bound0}, from which we can see that the cost actually determined by the effective number of data samples used in each iteration, i.e., $k\cdot b$. Since $K\propto\rho_C$ and $b\propto 1/\rho_C$, $\rho_C$ cancels out in $K\cdot b$. As a result, only $\rho_S$ makes a difference to the stability cost. This implies that \texttt{FATS} can achieve high client-level stability without compromising accuracy or communication efficiency if each client has enough local data samples. 
    
    \noindent (III) The second term scales as $O\left(\frac{E}{T}\right)$, which is controlled by the ratio of $E$ and $T$. To make this term diminish with $T\to\infty$, one can set $E=O(T^{1-\alpha})$ with $0<\alpha\le 1$. This way, the second term scale as $O(T^{-\alpha})$, and when $T$ is large enough, the entire bound will converge to the stability cost only. This actually reflects a trade-off between the convergence rate and the communication costs: larger $\alpha$ means faster convergence but also leads to more communication rounds. In some extreme cases, one can even set $E=c\cdot T$ for some constant $c<\frac{1}{2}\sqrt{\frac{\Gamma}{\lambda}}$. However, in this case, to retain the best possible convergence error of $O\left(\frac{1}{\sqrt{\rho_S MN}}\right)$, we will need an additional mild restriction on the client number $M$ so that $M\le O(N)$. This shows that \texttt{FATS} can achieve a fast convergence rate with a small number of communication rounds under certain conditions.
\end{remark}

With the perceptions above, we give in the following corollary the specific convergence error for different choices of $E$.

\begin{corollary}\label{coro-convergence}
    Under Assumption~\ref{asp-smoothness},~\ref{asp-localvar} and~\ref{asp-heterogeneity}, for any $0<\alpha\le 1$ if we set $\eta=\frac{1}{L\sqrt{\Gamma}T}$, $E=T^{1-\alpha}$, and $T\ge\max\left\{\frac{2\lambda}{\sqrt{\Gamma}}, \left(2\sqrt{\frac{\lambda}{\Gamma}}\right)^{\frac{1}{\alpha}}, \left(\rho_C M\right)^{\frac{1}{\alpha}}\right\}$,
    then we have
    \begin{flalign*}
        &\frac{1}{T}\sum_{t=1}^{T}\mathbb{E}[\|\nabla F(\theta^{(t-1)})\|_2^2]\le \frac{4G\sqrt{L(F(\theta^{(0)})-F^*)}}{\sqrt{\rho_S MN}}=O\left(\frac{1}{\sqrt{\rho_S MN}}\right).
    \end{flalign*}
    If we set $E=c\cdot T$ for some constant $c<\frac{1}{2}\sqrt{\frac{\Gamma}{\lambda}}$, $T\ge\frac{2\lambda}{\sqrt{\Gamma}}$ and $M<\frac{4\lambda^2L(F(\theta^{(0)})-F^*)\rho_S}{G^2\rho_C^2}N$, then we have
    \begin{flalign*}
        \frac{1}{T}\sum_{t=1}^{T}\mathbb{E}[\|\nabla F(\theta^{(t-1)})\|_2^2]\le\frac{4G\sqrt{L(F(\theta^{(0)})-F^*)}}{\sqrt{\rho_S MN}}=O\left(\frac{1}{\sqrt{\rho_S MN}}\right).
    \end{flalign*}
\end{corollary}

\begin{remark}
    The convergence error of \texttt{FATS} is dominated by the stability cost of $O\left(\frac{1}{\sqrt{\rho_S MN}}\right)$, where $\rho_S$ is closely related to the unlearning efficiency. Specifically, the smaller $\rho_S$ is, the more stable the algorithm is and the more efficient the unlearning process. In this way, $\rho_S$ characterizes the trade-off between the unlearning efficiency and learning accuracy. If we disregard unlearning efficiency and accept retraining computation for every sample removal request, then we can make $\rho_S>1$ arbitrarily large to obtain, as expected, arbitrarily small convergence error. However, the intriguing case is when we make $\rho_S<1$: in this case, we achieve a faster unlearning time and still a non-trivial accuracy, up to $\rho_S>O(\frac{1}{MN})$. This implies that we can attain a substantial decrease in communication and computation costs for federated unlearning without compromising too much accuracy for learning.
\end{remark}

\begin{remark}[Utility of unlearned models]
    Previously, we have shown a convergence error of $O\left(\frac{1}{\sqrt{\rho_S MN}}\right)$ for our FL algorithm \texttt{FATS}. This error bound reflects the accuracy of the original model trained on the full dataset. However, when we perform federated unlearning, we remove some data samples or clients from the dataset, which may affect the utility of the unlearned models. We now discuss the utility of the unlearned models and show that they preserve the same error bound as the original model under certain conditions. Since the error bound is mainly dependent on the total number of data samples $MN$, as long as the number of remaining data samples after deletion is still $O(MN)$, the convergence error bound of $O\left(\frac{1}{\sqrt{\rho_S MN}}\right)$ will still hold for the unlearned models. This means that the unlearned models have almost the same accuracy performance as the original model while satisfying the exact unlearning provability criterion. This conclusion will also be supported by our experiments. 
\end{remark}

\subsection{Unlearning Time and Space Overheads}

\subsubsection{Time Overheads.} We analyze the computational efficiency of \texttt{FATS}, which is a crucial aspect of federated unlearning, as it influences the scalability and responsiveness of the unlearning process. The total execution time of the unlearning algorithm consists of the time for verification and the time for re-computation. The most optimal way for verification is to maintain a dictionary of samples/clients to the earliest iteration/round that each sample/client participated in. This way, it requires $O(1)$ time lookup for every unlearning request. The time for each re-computation is bounded by the training time. According to Proposition~6 in \cite{ullah2021machine}, for a coupling-based unlearning algorithm with acceptance probability at least $1-\delta$ and for $w$ unlearning requests in general, the expected number of times re-computation is invoked is at most $4w\delta$. \texttt{FATS-SU} and \texttt{FATS-CU} re-compute with probability of $\min\{1, \rho_S\}$ and $\min\{1, \rho_C\}$, respectively. Hence, we can derive the expected number of re-computations for w sample-level or client-level unlearning requests as $4w\min\{1, \rho_S\}$ and $4w\min\{1, \rho_C\}$, respectively. Based on the analysis above, we obtain the expected running time for our framework as follows.

\begin{theorem}\label{th-time}
    For $w$ sample-level unlearning requests, the expected unlearning running time  is $O(\max\{\min\{\rho_S, 1\}w\times \text{Training time},w\})$. Similarly, for $w$ client-level unlearning requests, the expected running time is $O(\max\{\min\{\rho_C, 1\}w\times \text{Training time},w\})$.
\end{theorem}

\subsubsection{Space Overheads.} 
We require the server and the local devices to store some intermediate states of the training process for \texttt{FATS} as in Algorithm~1. We enumerate the parameters to be stored on the server and on each local device. The server stores the subsets of clients in each communication round (i.e., $\mathcal{P}^{(t)}$) and the global models (i.e., $\theta^{(t)}$). For every training iteration $t$ that client $k$ is selected, client $k$ preserves the iteration index (i.e., $t$), local mini-batches (i.e., $\mathcal{B}_k^{(t)}$) and local models (i.e., $\theta_k^{(t)}$). We conduct a space complexity analysis of our algorithms. Each local device stores all the mini-batches and local models that it utilizes for every training iteration, which consumes up to $O(T\cdot b)$ and $O(T\cdot d)$ words respectively via dictionaries that associate iteration index with mini-batch samples and local models. The space complexity for each device is $O(T\cdot\max\{b, d\})=O\left(T\cdot\max\left\{\frac{\rho_S N}{\rho_C E}, d\right\}\right)$. The server stores the involved clients and the global models in each communication round, which occupies $O(R\cdot K)$ and $O(R\cdot d)$ words respectively via dictionaries that link round index with client sub-sets and global models. The space complexity for the server is $O(R\cdot\max\{K,d\})=O\left(\max\left\{\rho_C M, \frac{Td}{E}\right\}\right)$.

We remark that our unlearning algorithms employ the stored local models, mini-batch samples, and client sub-sets only for presentation convenience. We can easily devise a simple but efficient implementation of our unlearning algorithms, which does not store any local models, mini-batch samples, or client sub-sets, but has the same unlearning time complexity as in Theorem~\ref{th-time}. In Theorem~\ref{th-time}, we bound the re-computation time by the full re-computation time, i.e., training time. This implies that the unlearning time bound holds even if we do full re-computation every time the discrepancy arises. The unlearning algorithms can be modified as: instead of retraining from iteration $t_0$ where the target sample or client is involved, we do full retraining from the beginning. With this, each local device stores a vector of length $N$ to indicate the involvement of each local sample, and the server stores a vector of length $M$ to indicate the participation of each client, which takes $O(N)$ and $O(M)$ bits respectively. Both the server and each device store a $d$-dimensional model, which costs $O(d)$ words. The space complexities of each device and the server are $O(N+d)$ and $O(M+d)$ words, which are independent of $T$ and acceptable even when $T$ is large.

\section{Experiments}\label{sec-exp}
To demonstrate the advantages of our proposed framework in terms of performance and scalability of learning/unlearning, we conduct comprehensive experiments on 6 federated learning benchmark datasets and present empirical evidence of its superiority. Specifically, we compare our framework with state-of-the-art methods for federated unlearning and evaluate the accuracy, communication cost, computation cost, and unlearning effectiveness of different methods. We also investigate the impact of different stability parameters on the performance of our framework.

\begin{figure*}[htbp]
     \centering
          \begin{subfigure}[b]{0.3\textwidth}
         \centering
         \includegraphics[width=\textwidth]{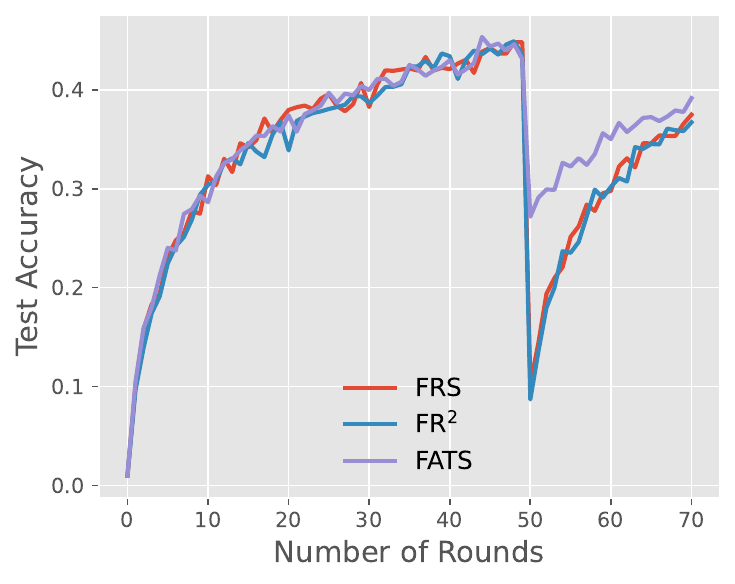}
         \caption{Cifar-100}
     \end{subfigure}
     \hfill
     \begin{subfigure}[b]{0.3\textwidth}
         \centering
         \includegraphics[width=\textwidth]{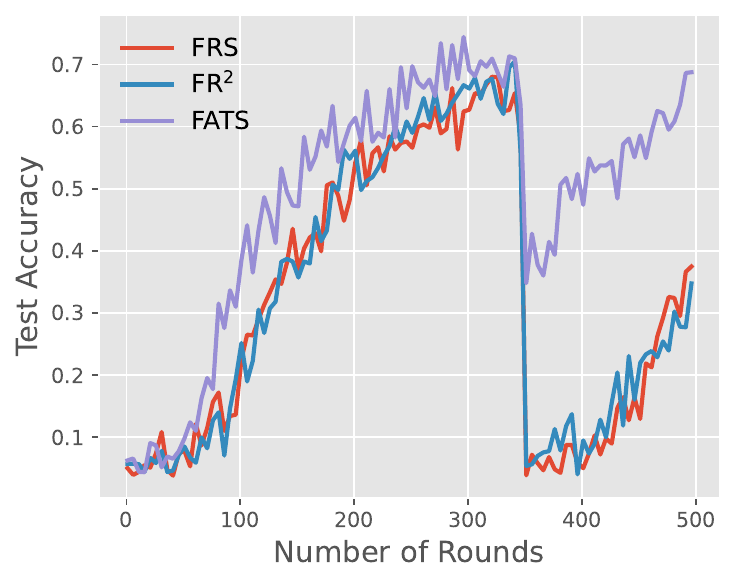}
         \caption{FEMNIST}
     \end{subfigure}
     \hfill
     \begin{subfigure}[b]{0.3\textwidth}
         \centering
         \includegraphics[width=\textwidth]{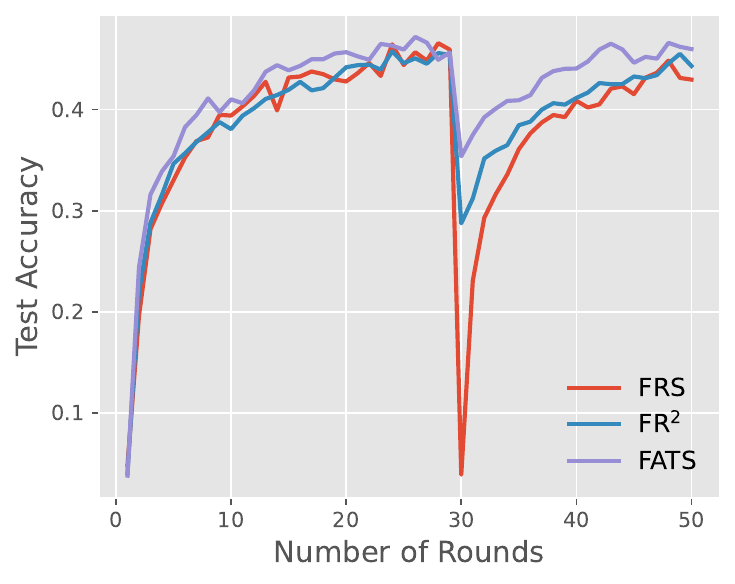}
         \caption{Shakespeare}
     \end{subfigure}
     \vfill
     \begin{subfigure}[b]{0.3\textwidth}
         \centering
         \includegraphics[width=\textwidth]{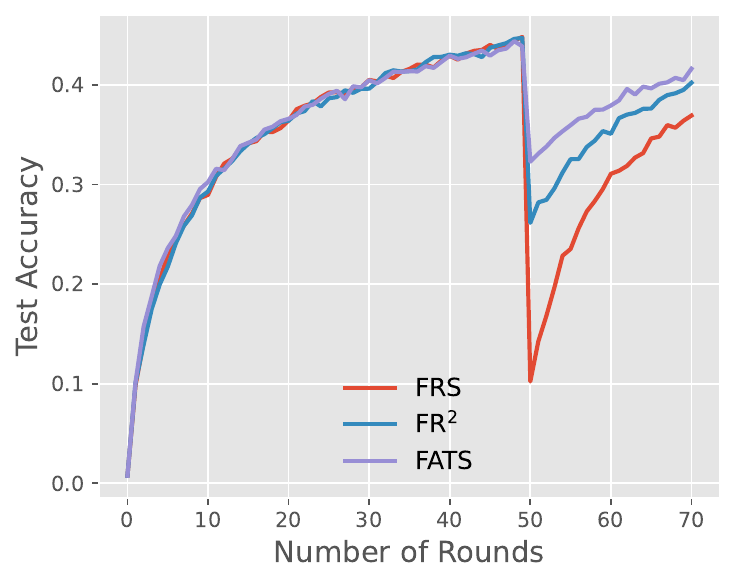}
         \caption{Cifar-100}
     \end{subfigure}
     \hfill
     \begin{subfigure}[b]{0.3\textwidth}
         \centering
         \includegraphics[width=\textwidth]{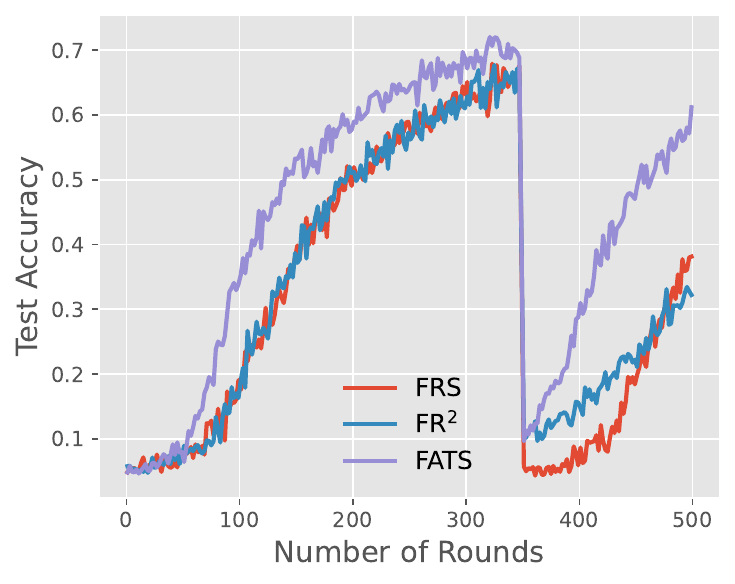}
         \caption{FEMNIST}
     \end{subfigure}
     \hfill
     \begin{subfigure}[b]{0.3\textwidth}
         \centering
         \includegraphics[width=\textwidth]{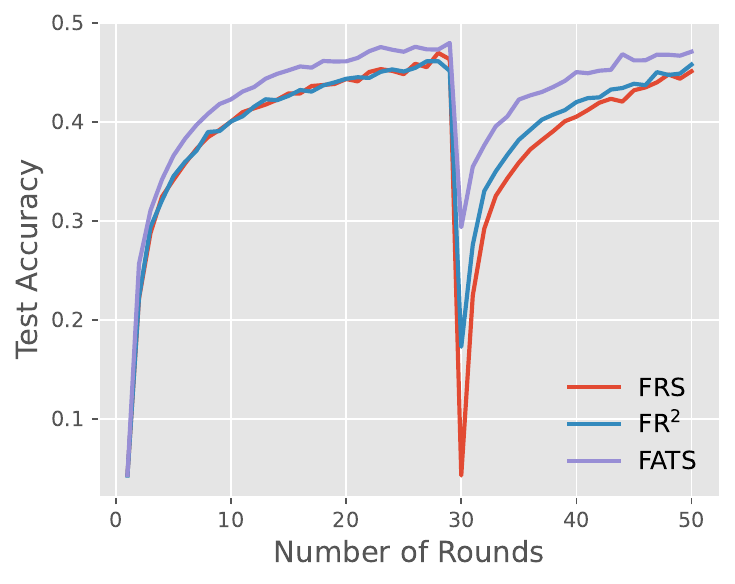}
         \caption{Shakespeare}
     \end{subfigure}
        \caption{Comparison of the test accuracy of different methods and their changes after unlearning on Cifar-100, FEMNIST, and Shakespeare. \textit{Top:} sample-level unlearning. \textit{Bottom:} client-level unlearning.}
        \label{fig:performance}
\end{figure*}

\subsection{Experimental Settings}
\subsubsection{Datasets}
We adopt several widely-used benchmark datasets. These datasets can be categorized into two categories: \textit{simulated federated datasets} and \textit{real federated datasets}. For simulated federated datasets, we manually partition centralized datasets into the federated setting. For real federated datasets, we use datasets from the well-known FL benchmark LEAF~\cite{LEAF} whose datasets are originally built for real-world federated settings.

\textbf{Simulated federated datasets.} We use MNIST~\cite{lecun1998gradient}, Fashion-MNIST (FashionM)~\cite{xiao2017fashion}, Cifar-10 and Cifar-100~\cite{krizhevsky2009learning} and apply the Label-based Dirichlet Partition (LDA)~\cite{LDA19} on them to simulate the label-based non-i.i.d. data distribution. These datasets are widely used for image classification tasks in FL, e.g.,~\cite{mcmahan2017communication}. Each dataset is partitioned into 100 clients using the LDA, assigning the partition of samples to clients by $p \sim \operatorname{Dir}(\beta)$, where $\operatorname{Dir}$ is the probability density function of the Dirichlet distribution and $\beta$ is a positive real parameter. Smaller $\beta$ leads to a higher heterogeneity level of partition. Unless otherwise stated, we set $\beta$ to $0.5$ by default.

\textbf{Real federated datasets.} We use FEMNIST and Shakespeare (Shakes) from LEAF and they cover both vision and language tasks. We sample the full-sized datasets and partition each client's samples into training/test groups as per \cite{LEAF} except that we set the minimum number of samples per client to $100$.  

For more details of the datasets, see Table~\ref{tab:datasets} in Appendix~\ref{sec:dataset}.

\subsubsection{Models and Hyperparameters} 

We adopt different models for different FL tasks, based on the complexity and modality of the data. Specifically, for image classification tasks, we use the CNN model for MNIST, Fashion-MNIST, FEMNIST, and the VGG16 pre-trained on ImageNet1K for Cifar-10 and Cifar-100. 
The learning rate is set to be $0.001$ for both models. For the natural language generation task on Shakespeare, we use the same model architectures as reported in ~\cite{LEAF}: the model first maps each character to an embedding of dimension 8 before passing it through an LSTM of two layers of $256$ units each. The LSTM emits an output embedding, which is scored against all items of the vocabulary via a dot product followed by a softmax. 
We use a sequence length of $80$ for LSTM and the learning rate of LSTM is set as $0.8$. All other hyperparameters are set to the values reported in Table~\ref{tab:datasets} for the respective datasets. 

\begin{figure}[ht]
     \centering
     \begin{subfigure}[b]{0.82\linewidth}
         \centering
         \includegraphics[width=\textwidth]{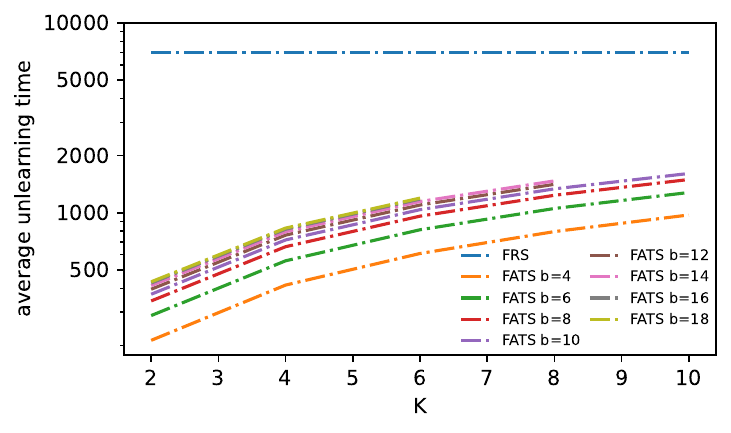}
         \caption{FEMNIST: Sample Unlearning Case}
     \end{subfigure}
     \vfill
     \begin{subfigure}[b]{0.82\linewidth}
         \centering
         \includegraphics[width=\textwidth]{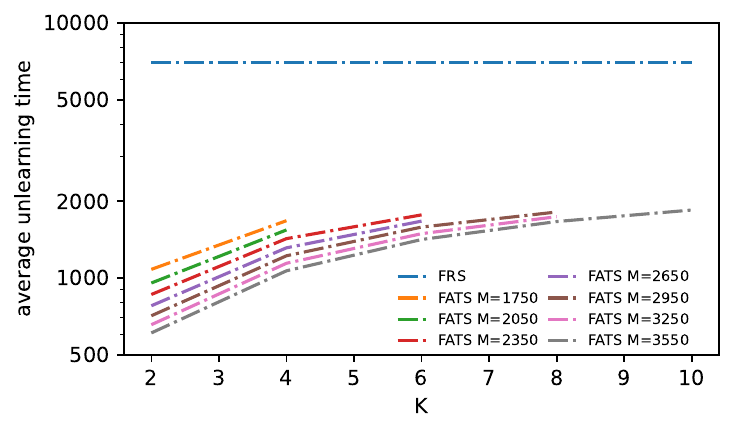}
         \caption{FEMNIST: Client Unlearning Case}
     \end{subfigure}
     \hfill
    \caption{Unlearning Efficiency of \texttt{FATS} compared with \texttt{FRS}.}
    \label{fig:efficiency}
\end{figure}

\begin{figure}
     \centering
     \begin{subfigure}[b]{0.82\linewidth}
         \centering
         \includegraphics[width=\textwidth]{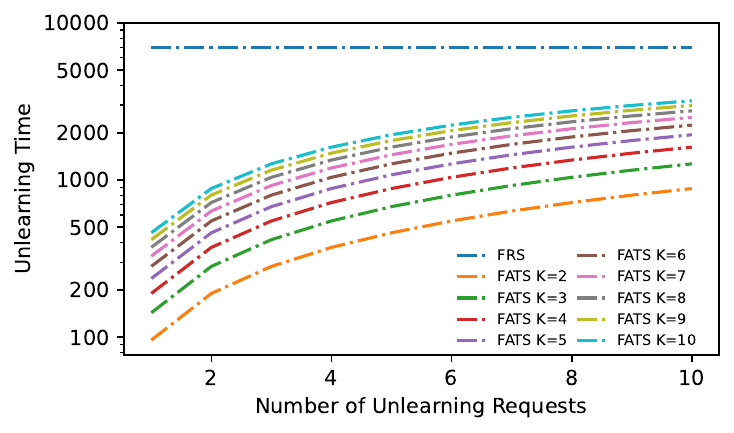}
         \caption{FEMNIST}
     \end{subfigure}
     \hfill
     \begin{subfigure}[b]{0.82\linewidth}
         \centering
         \includegraphics[width=\textwidth]{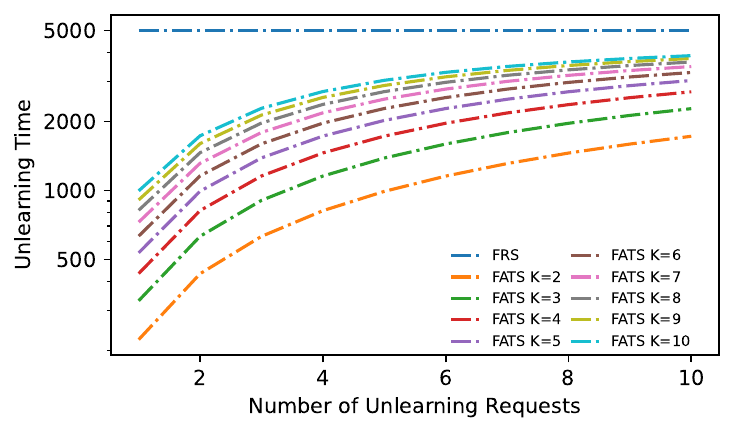}
         \caption{Shakespeare}
     \end{subfigure}
     \caption{Impacts of the number of unlearning requests on unlearning efficiency.}
        \label{fig:Unumber}
\end{figure}

\begin{figure}
     \centering
     \begin{subfigure}[b]{0.8\linewidth}
         \centering
         \includegraphics[width=\textwidth]{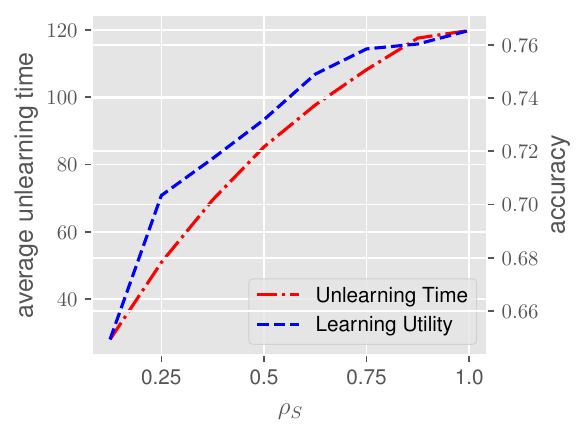}
         \caption{Fashion-MNIST: varying $\rho_S$}
     \end{subfigure}
     \hfill
     \begin{subfigure}[b]{0.8\linewidth}
         \centering
         \includegraphics[width=\textwidth]{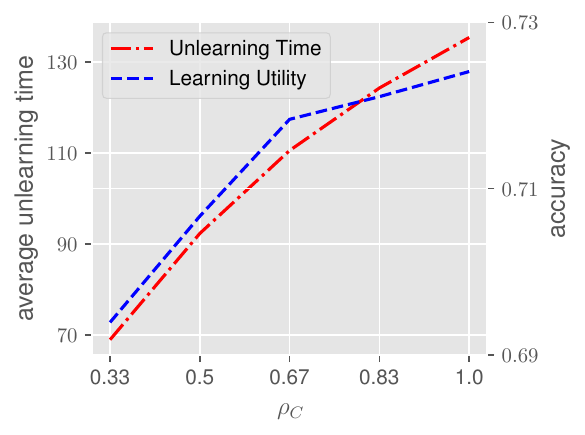}
         \caption{Fashion-MNIST: varying $\rho_C$}
     \end{subfigure}
        \caption{Impacts of stability parameters on learning utility and unlearning efficiency.}
        \label{fig:UvsE}
\end{figure}

\subsubsection{Evaluation Metrics} 

We use the following metrics to assess the learning and unlearning performance of different methods on various datasets and tasks.

\textbf{Test accuracy.} This metric quantifies the accuracy of the global model for different tasks on the test data of various datasets. For image datasets, we assess the classification accuracy of the model on the test data, which means the ratio of correct predictions to the total number of predictions. For Shakespeare dataset, we gauge the top-1 accuracy of the model for predicting the next word in a sequence, which means the fraction of the model answer (the one with highest probability) that matches the actual words in the test data. Higher accuracy indicates better performance.

\textbf{Unlearning time}. This metric measures the number of time steps required by an unlearning algorithm to unlearn target data samples or clients. It reflects the computation/communication efficiency and scalability of the unlearning algorithm. Fewer unlearning time means higher efficiency.

\textbf{MIA accuracy and precision}. These two metrics measure the vulnerability of the global model under the membership inference attack (MIA)~\cite{ShokriSSS17}. MIA aims to infer whether a given target data sample was used to train the model or not. Thus, the performance of MIA reflects the amount of information that remains in the unlearned global model, which indicates the unlearning efficacy. Accuracy and precision are two complementary metrics for MIA to assess how well the attack can infer the membership status of a given data point. Accuracy is the fraction of correct predictions over all predictions, while precision is the fraction of correct positive predictions over all positive predictions. In other words, accuracy measures how often the attack is right, while precision measures how confident the attack is when it predicts a positive membership. When applying MIA to unlearned models, the closer the MIA accuracy or precision is to 50\% (which intuitively represents a random guess), the higher the unlearning efficacy.

For the test accuracy and unlearning time, we will report the averaged results over 5 independent runs. For MIA accuracy and precision, we will perform MIA 100 times and report the average performance as well as the standard deviation.

\subsubsection{Baseline Methods} 

To show the advantages of our framework, we compare it with two existing methods: Federated Retraining from Scratch (\texttt{FRS}) and Federated Rapid Retraining (\texttt{FR}$^2$)~\cite{liu2022right}. For these two methods, we use the most widely used FL algorithm, federated averaging (\texttt{FedAvg})~\cite{mcmahan2017communication}, to train the global FL model. The unlearning strategy of these two methods is described as follows.

\textbf{\texttt{FRS}}. This method retrains the FL model from scratch on the remaining data after removing target samples or clients. This is the simplest but most time and communication intensive method.

\textbf{$\texttt{FR}^2$}. This method employs a diagonal approximation of the Fisher Information Matrix to achieve rapid retraining and introduces the momentum technique to enhance model utility. This is a state-of-the-art federated unlearning framework, applicable to sample-level and client-level unlearning.

We implement all methods based on Flower~\cite{FLOWER}, a scalable and efficient open-source FL framework.

\begin{table*}[tbp]
\centering
\caption{Membership Inference Attack (MIA) Results on Six Datasets.}
\begin{tabular}{l|ccc|ccc} 
\hline\hline
\multirow{2}{*}{Datasets} & \multicolumn{3}{c|}{Accuracy}                                                                                                                              & \multicolumn{3}{c}{Precision}                                                                                                                              \\ 
\cdashline{2-7}[1pt/1pt]
                          & \texttt{FRS}                                               & \texttt{FR}$^2$                                            & \texttt{FATS}                                               & \texttt{FRS}                                               & \texttt{FR}$^2$                                            & \texttt{FATS}                                               \\ 
\hline\hline
MNIST                     & 50.10$\pm$ 0.25\%                    & 49.47$\pm$ 0.42\%                    & 50.06$\pm$ 1.40\%                     & 50.55$\pm$ 0.13\%                    & 49.70$\pm$ 0.24\%                    & 50.30$\pm$ 0.78\%                     \\
FMNIST                    & 47.85$\pm$ 1.93\%                    & 48.47$\pm$ 5.20\%                    & 48.29$\pm$ 5.28\%                     & 49.13$\pm$ 1.55\%                    & 48.91$\pm$ 4.99\%                    & 49.48$\pm$ 4.40\%                     \\
Cifar-10                   & 49.04$\pm$4.03\%                     & 51.23$\pm$2.25\%                     & 49.15$\pm$1.91\%                      & 49.53$\pm$11.34\%                    & 54.17$\pm$7.48\%                     & 49.20$\pm$0.38\%                      \\
Cifar-100                  & \multicolumn{1}{l}{49.34$\pm$0.53\%} & \multicolumn{1}{l}{47.89$\pm$1.53\%} & \multicolumn{1}{l|}{48.61$\pm$1.29\%} & \multicolumn{1}{l}{49.20$\pm$0.61\%} & \multicolumn{1}{l}{48.03$\pm$1.39\%} & \multicolumn{1}{l}{48.90$\pm$0.99\%}  \\
FEMNIST                   & 50.14$\pm$0.23\%                     & 50.00$\pm$0.00\%                     & 53.30$\pm$2.34\%                      & 55.01$\pm$17.22\%                    & 33.20$\pm$42.12\%                    & 55.64$\pm$4.72\%                      \\
Shakes                    & 49.68$\pm$0.26\%                     & 49.44$\pm$0.04\%                     & 49.35$\pm$0.02\%                      & 49.30$\pm$0.51\%                     & 48.87$\pm$0.13\%                     & 49.04$\pm$0.28\%                      \\
\hline\hline
\end{tabular}
\label{tab2}
\end{table*}

\subsection{Experimental Results}

\subsubsection{Performance Evaluation}\label{sec:performance}

We compare the learning performance and the unlearning capability of our proposed method with two baseline methods, \texttt{FRS} and \texttt{FR}2, on all 6 datasets. Our main focus in this experiment is to examine how these methods can achieve FL training, handle unlearning requests, and recover from data deletion. To this end, we first train a global FL model for all these methods, then we issue some unlearning requests to them and let them update the model accordingly. We measure and present the model test accuracy throughout the whole process. Due to space limit, here we report the results on Cifar-100, FEMIST and Shakespeare in Figure~\ref{fig:performance}. Other results are included in Appendix~\ref{sec:add_Performance}. We will also study the streaming unlearning setting in Appendix~\ref{sec:add_streaming}.

Regarding the unlearning request issue time, we vary the issue iterations for different datasets according to their convergence speed. In fact, there is no clear criterion to decide when an FL model is fully trained, and unlearning requests can be issued at any time during the FL training process. If an unlearning request is issued in the initial stage of FL training, even the naive retraining method would be relatively cheap. Therefore, in this experiment, we choose an issue iteration for each dataset such that the test accuracy has reached a stable level. At such an iteration, the FL model has been adequately trained and the performance gap between different unlearning methods is evident. For the unlearning requests, we consider both sample-level and client-level unlearning scenarios. Specifically, we randomly select 10 samples/clients for MNIST and FEMIST and 5 samples/clients for the other datasets to be unlearned simultaneously. The reason why we issue a batch of unlearning requests is that, if we only issue one request, re-computation may not be triggered in \texttt{FATS}, and then we will not be able to compare how these methods cope with sample or client deletion.

For the learning process, we can see that \texttt{FATS} achieves comparable or even superior (e.g., for FEMNIST and Shakespeare) test accuracy than the baselines before unlearning over all the datasets. This indicates that \texttt{FATS} can learn as effectively as the classical FL algorithm \texttt{FedAvg} while achieving TV-stability at the same time.

For the unlearning process, we can see that our method consistently outperforms the baselines across all datasets. \texttt{FRS} suffers from a significant drop in accuracy after data deletion, which incurs more rounds to recover the model utility. This shows that \texttt{FRS} is inefficient and costly for unlearning, as it requires retraining the model from scratch on the updated dataset. \texttt{FR}$^2$ has difficulty on converge and requires more communication rounds to achieve stable performance, as shown by the large fluctuations on MNIST, Fashion-MNIST, and FEMNIST. This shows that \texttt{FR}$^2$ is not robust and unreliable for unlearning, as it relies on a diagonal approximation of the Fisher Information Matrix that may not capture the true Hessian information. In contrast, our framework achieves rapid and effective unlearning with minimal loss in accuracy and communication cost. The accuracy drop of \texttt{FATS} is significantly smaller than that of all baselines over all the datasets. This shows that our method can unlearn more quickly and accurately without affecting the model utility too much. This demonstrates the superiority of our method over the existing methods for federated unlearning.

In summary, \texttt{FATS} achieves effective learning and unlearning performance in the federated setting. It learns as fast as \texttt{FedAvg} and achieves comparable or better accuracy before unlearning. It unlearns more rapidly and accurately than the baselines and preserves more model utility after unlearning.

\subsubsection{Unlearning Efficiency}\label{sec:unlearning_efficiency}
The unlearning efficiency of our framework is influenced by the stability parameters $\rho_S$ and $\rho_C$ for sample-level and client-level unlearning, respectively. These parameters are related to the hyperparameters $K$, $T$, $E$, $M$, $b$, and $N$. Specifically, $\rho_C=(K\cdot T)/(E\cdot M)$ and $\rho_S=(b\cdot K\cdot T)/(N\cdot M)$. We aim to assess the unlearning efficiency of our framework and contrast it with the baseline methods under different settings of stability parameters $\rho_S$ and $\rho_C$. We manipulate $\rho_S$ and $\rho_C$ by adjusting the values of the hyperparameters. Specifically, we keep $T$ and $E$ constant as shown in Table~\ref{tab:datasets} in this experiment and only vary $M$, $K$ and $b$. We test the unlearning efficiency of \texttt{FATS} on the FL benchmark datasets FEMNIST and Shakespeare. We present the results on FEMNIST in Figure~\ref{fig:efficiency}. The results on Shakespeare can be found in Appendix~\ref{sec:add_efficiency}. 

For sample-level unlearning, we fix the hyperparameters $T$, and $M$, and vary $K$ for each chosen $b$, which changes the sample unlearning ratio $\rho_S$ accordingly. Since $\rho_S$ is proportional to $K$ when $M$, $N$, $T$, and $b$ are fixed, a larger $K$ corresponds to a higher $\rho_S$. The results are shown in the first row of Figure~\ref{fig:efficiency}. Each line therein ends at the largest $K$ such that $\rho_S$ reaches $1$. We can see that \texttt{FATS} requires fewer rounds to fully recover model utility for sample unlearning, compared to the baseline \texttt{FRS}.

For client-level unlearning, we fix the hyperparameters $T$ and $E$, and vary $K$ for each chosen $M$, which changes the client unlearning ratio $\rho_C$ accordingly. Since $\rho_C$ is proportional to $K$ when $M$, $T$, and $E$ are fixed, a larger $K$ corresponds to a higher $\rho_C$. The results are shown in the second row of Figure~\ref{fig:efficiency}. Each line therein ends at the largest $K$ such that $\rho_C$ reaches $1$. We find that \texttt{FATS} takes more time for larger values of $K$ as expected, but its largest average unlearning time is still less than half of the time taken by \texttt{FRS}.

 In summary, our proposed federated unlearning framework demonstrates superior efficiency over baseline approaches on large real-world federated datasets. It accomplishes unlearning with less time and rounds while restoring model accuracy. These results highlight the effectiveness of our proposed method for efficient federated unlearning.

\subsubsection{Impact of the Number of Unlearning Requests}

As stated in Theorem~\ref{th-time}, the time taken to unlearn data is greatly affected by the number of unlearning requests. In this experiment, we examine how the number of unlearning requests influences the unlearning efficiency in practice. We conduct experiments on FEMNIST and Shakespeare. For FEMNIST, we use $M=3550$ clients and for Shakespeare, we use $M=630$ clients. For both datasets, we consider the settings with K varying from $2$ to $10$. Since different K corresponds to different stability parameter $\rho_C$, we also examine whether the influence of unlearning request number holds consistently under different stability parameters. We issue $1-10$ client unlearning requests respectively for each setting and report the average unlearning time in Figure~3. For comparison, we choose \texttt{FRS} as the baseline method. As we can see from Figure~3, when $\rho_C$ is fixed, the unlearning time is positively correlated with the number of unlearning requests. Moreover, since larger $K$ corresponds to larger $\rho_C$, we can also see that, the unlearning time is also positively correlated with the stability parameter when the unlearning request number is fixed. We can also observe that with some appropriate $K$, the unlearning time can still be significantly smaller than it of FRS even if there are multiple unlearning requests. All these experimental results corroborate our theoretical results.

\subsubsection{Utility v.s. Efficiency}
To further analyze the trade-off between learning utility and unlearning efficiency, we show the average unlearning time and accuracy of \texttt{FATS} as a function of $\rho_C$ and $\rho_S$ on MNIST and Fashion-MNIST. Here we present the results on Fashion-MNIST in Figure~\ref{fig:UvsE}. The results on MNIST can be found in Appendix~\ref{sec:add_UvE}. The first row of Figure~\ref{fig:UvsE} shows how the average unlearning time and accuracy of FATS change as $\rho_S$ increases from 0.125 to 1. We can see that the accuracy curves rise rapidly at first, then slowly level off. The unlearning time curves exhibit a similar trend, decreasing sharply initially before plateauing as $\rho_S$ approaches 1. The second row of Figure~\ref{fig:UvsE} shows how the average unlearning time and accuracy of FATS change as $\rho_C$ increases from 0.2 to 1 for MNIST and from 0.33 to 1 for Fashion-MNIST. We can see that the accuracy curves increase faster than the unlearning time curves at first, but then start to flatten out when $\rho_C$ exceeds 0.5, while the unlearning time curves keep increasing. This suggests an optimal trade-off point around $\rho_C=0.5$, where good learning utility is balanced with efficient client-level unlearning. Overall, we can observe that $\rho_S$ has a higher impact on the learning utility than $\rho_C$. This is also supported by the convergence analysis we conducted in Section~\ref{sec:convergence}.

\subsubsection{Membership Inference Attack}

We conduct \textit{Membership Inference Attack} (MIA)~\cite{ShokriSSS17} on the final unlearned global models trained by different federated unlearning methods. The MIA results are shown in Table~\ref{tab2}. Our method exhibits high unlearning effectiveness, as the attack gains no advantage over random guessing and cannot reliably discern membership in the training set. The performance of \texttt{FATS} and \texttt{FRS} is generally consistent. This suggests that our method indeed achieves \textit{exact} federated unlearning. We note that the state-of-the-art baseline \texttt{FR}$^2$ achieves very low MIA precision on the FEMNIST dataset, which indicates that \texttt{FR}$^2$ is not robust and unreliable for unlearning. In summary, the MIA results verify that our method indeed satisfies exact federated unlearning, and demonstrate the superiority of our method over the existing methods for federated unlearning.

\section{Conclusions}\label{sec-conclusions}
In this paper, we presented the first algorithmic framework for federated unlearning that attains communication efficiency and unlearning provability. The newly introduced notion of exact federated unlearning guarantees the total removal of a client’s or a sample’s effect on the global model. We devised a TV-stable FL algorithm \texttt{FATS}, which reduces the communication rounds by using local SGD with periodic averaging. We also developed matching unlearning algorithms for \texttt{FATS} to cope with both client-level and sample-level unlearning scenarios. Furthermore, we offered theoretical analysis to demonstrate that our learning and unlearning algorithms fulfill exact unlearning and achieve reasonable convergence errors for both the learned and unlearned models. Lastly, we verified our framework through comprehensive experiments on 6 benchmark datasets, and revealed that our framework outperforms existing baselines, while considerably lowering the communication and computation cost, and fully erasing the impact of a specific data sample or client from the global FL model.


\section*{Acknowledgments}
Dongxiao Yu, Xiuzhen Cheng and Youming Tao are supported in part by National Natural Science Foundation of China (NSFC) under Grant 62122042, Major Basic Research Program of Shandong Provincial Natural Science Foundation under Grant ZR2022ZD02, Fundamental Research Funds for the Central Universities under Grant 2022JC016. Di Wang and Cheng-long Wang are supported in part by the baseline funding BAS/1/1689-01-01, funding from the CRG grand URF/1/4663-01-01, REI/1/5232-01-01, REI/1/5332-01-01, FCC/1/1976-49-01 from CBRC of King Abdullah University of Science and Technology (KAUST). They are also supported by the funding RGC/3/4816-09-01 of the SDAIA-KAUST Center of Excellence in Data Science and Artificial Intelligence (SDAIA-KAUST AI). Miao Pan was supported in part by the US National Science Foundation under grant CNS-2107057 and CNS-2318664.



\balance
\bibliographystyle{ACM-Reference-Format}
\bibliography{sample}

\vspace{11pt}

\vfill

\newpage
\onecolumn
\appendix

\section{Additional Experiment Details and Results}
\subsection{Dataset Details for Experiments}\label{sec:dataset}

\begin{table}[htbp]
\centering
\caption{Datasets}
\begin{tabular}{lccccccr} 
\hline\hline
Dataset   & \# sample & M     & K  & R   & E   & b  & Model     \\ 
\hline\hline
MNIST     & 60,000    & 300   & 5  & 30 & 10  & 10 & CNN       \\
FashionM & 60,000    & 300   & 5  & 50 & 10  & 10 & CNN       \\
Cifar-10   & 60,000    & 600   & 5  & 50 & 10  & 10 & VGG16  \\ 
Cifar-100   & 60,000    & 600   & 10  & 50 & 10  & 10 & VGG16  \\ 
\hline
FEMNIST   & 811,586   & 3,556 & 5  & 350 & 20  & 10 & CNN       \\
Shakes      & 3,678,451 & 660   & 20 & 30  & 100 & 60 & LSTM      \\
\hline\hline
\end{tabular}
\label{tab:datasets}
\end{table}

\subsection{Additional Results for Performance Evaluation}\label{sec:add_Performance}

\begin{figure*}[htbp]
     \centering
     \begin{subfigure}[b]{0.33\textwidth}
         \centering
         \includegraphics[width=\textwidth]{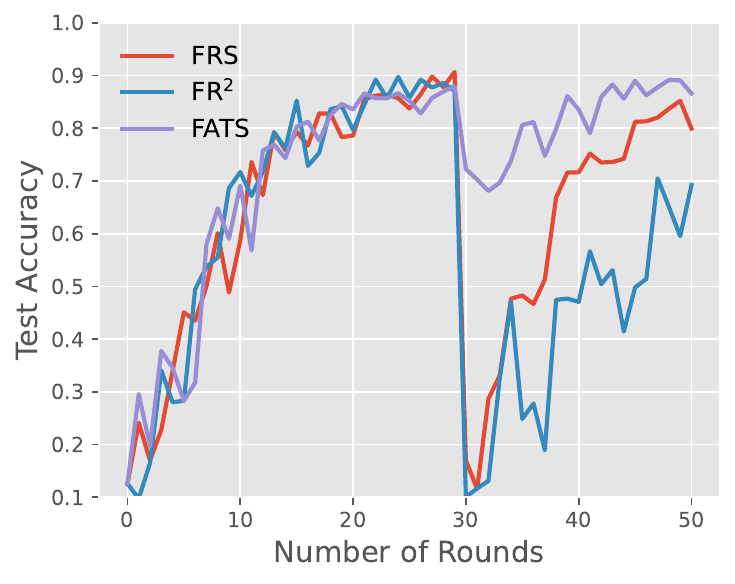}
         \caption{MNIST}
     \end{subfigure}
     \hfill
     \begin{subfigure}[b]{0.33\textwidth}
         \centering
         \includegraphics[width=\textwidth]{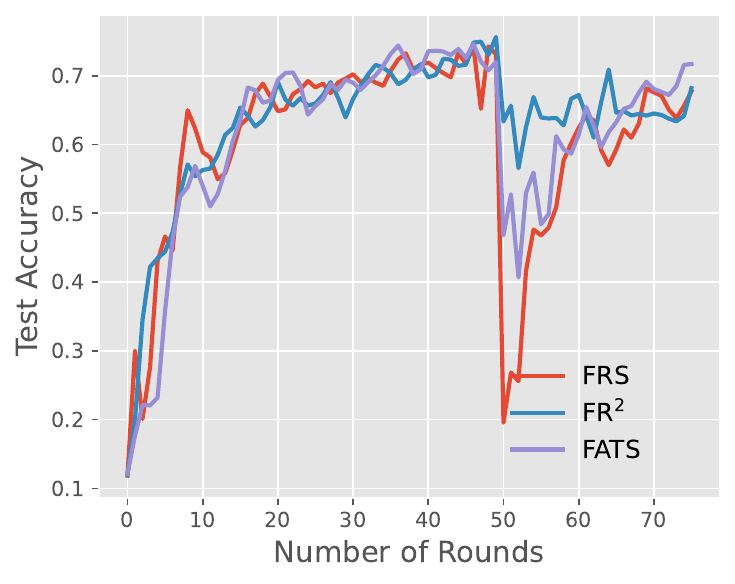}
         \caption{Fashion-MNIST}
     \end{subfigure}
     \hfill
     \begin{subfigure}[b]{0.33\textwidth}
         \centering
         \includegraphics[width=\textwidth]{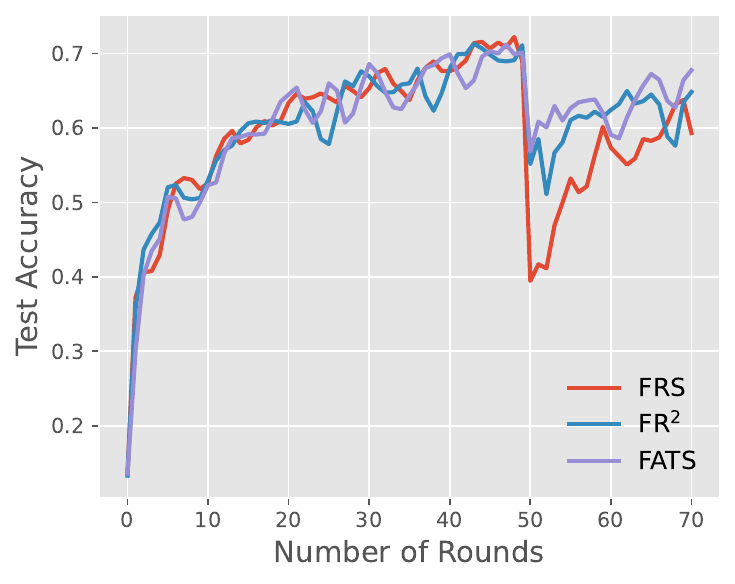}
         \caption{Cifar-10}
     \end{subfigure}
     \vfill
     \begin{subfigure}[b]{0.33\textwidth}
         \centering
         \includegraphics[width=\textwidth]{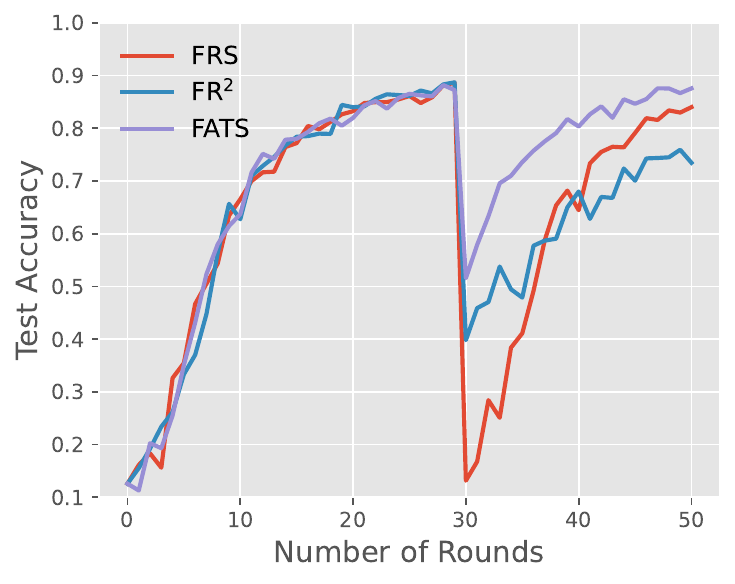}
         \caption{MNIST}
     \end{subfigure}
     \hfill
     \begin{subfigure}[b]{0.33\textwidth}
         \centering
         \includegraphics[width=\textwidth]{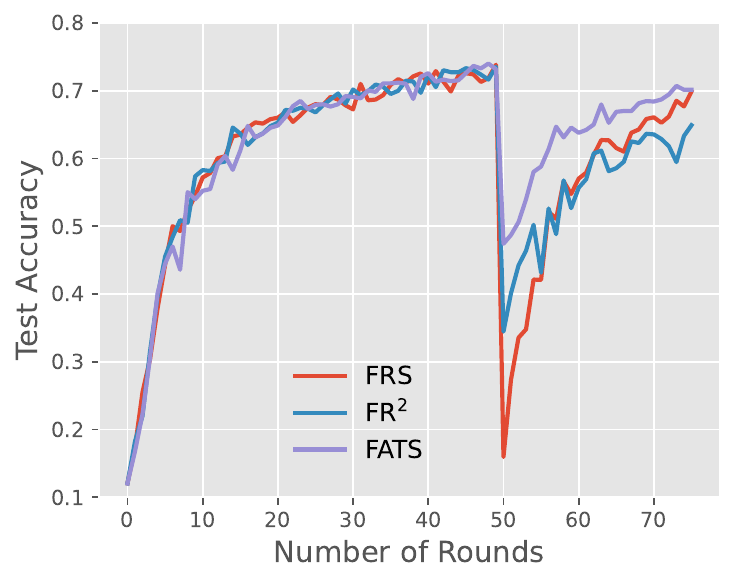}
         \caption{Fashion-MNIST}
     \end{subfigure}
     \hfill
     \begin{subfigure}[b]{0.33\textwidth}
         \centering
         \includegraphics[width=\textwidth]{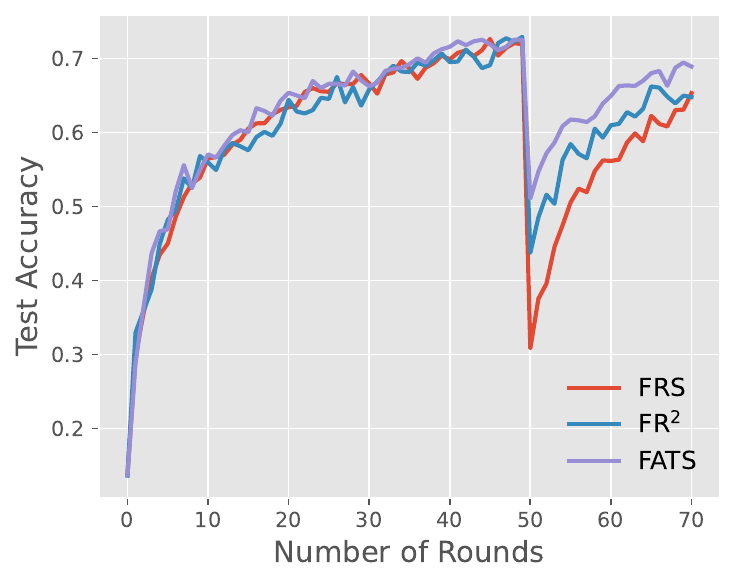}
         \caption{Cifar-10}
     \end{subfigure}
        \caption{Comparison of the test accuracy of different methods and their changes after conducting unlearning on MNIST, Fashion-MNIST, and Cifar-10. Top: sample-level unlearning. Bottom: client-level unlearning.}
        \label{fig:performance2}
\end{figure*}

We present the learning and unlearning performance evaluation results over MNIST, Fashion-MNIST, and Cifar-10 in Figure~\ref{fig:performance2}.

\begin{figure*}[ht]
     \centering
     \begin{subfigure}[b]{0.49\linewidth}
         \centering
         \includegraphics[width=\textwidth,height=0.6\textwidth]{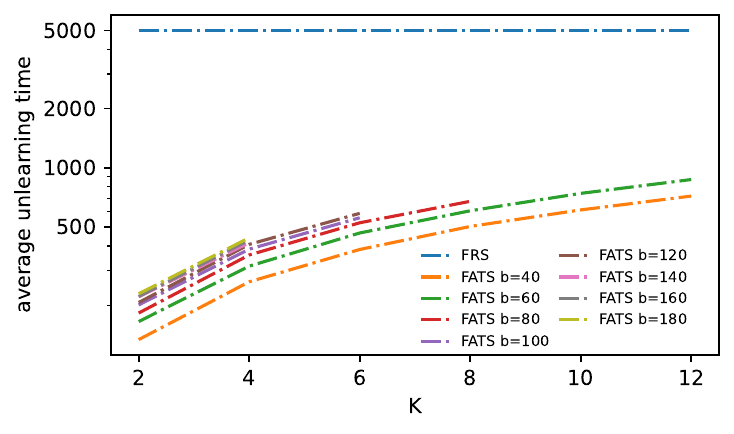}
         \caption{Shakespeare: Sample Unlearning}
     \end{subfigure}
     \hfill
     \begin{subfigure}[b]{0.49\linewidth}
         \centering
         \includegraphics[width=\textwidth]{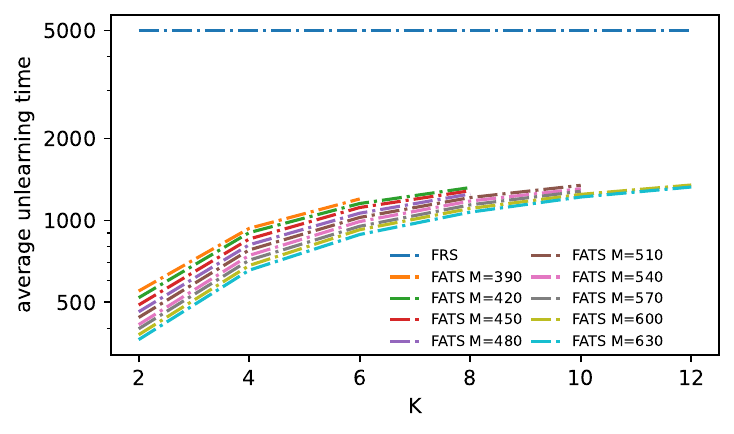}
         \caption{Shakespeare: Client Unlearning}
     \end{subfigure}
     \caption{Unlearning Efficiency of \texttt{FATS} compared with \texttt{FRS} on Shakespeare.}
     \label{fig:efficiency2}
\end{figure*}

\begin{figure}
     \centering
     \begin{subfigure}[b]{0.49\linewidth}
         \centering
         \includegraphics[width=\textwidth]{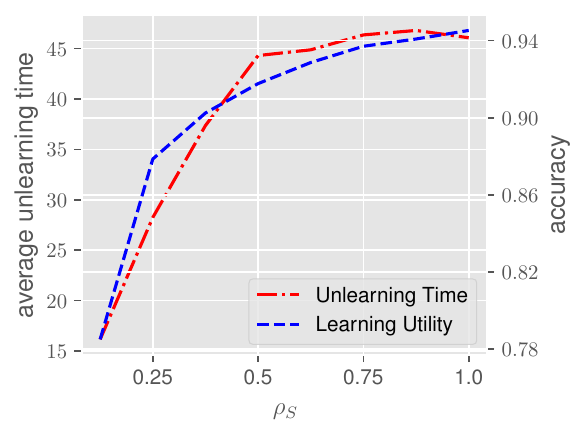}
         \caption{MNIST: Sampling Unlearning}
     \end{subfigure}
     \hfill
     \begin{subfigure}[b]{0.49\linewidth}
         \centering
         \includegraphics[width=\textwidth]{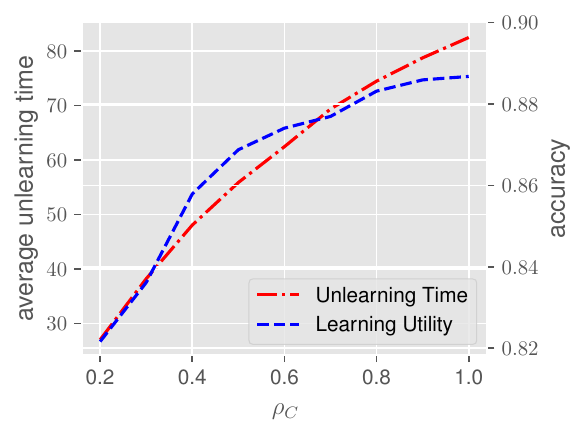}
         \caption{MNIST: Client Unlearning}
     \end{subfigure}
        \caption{Impacts of stability parameters on learning utility and unlearning efficiency}
        \label{fig:UvsE2}
\end{figure}

\begin{figure*}[ht]
     \centering
     \begin{subfigure}[b]{0.49\linewidth}
         \centering         
         \includegraphics[width=\textwidth]{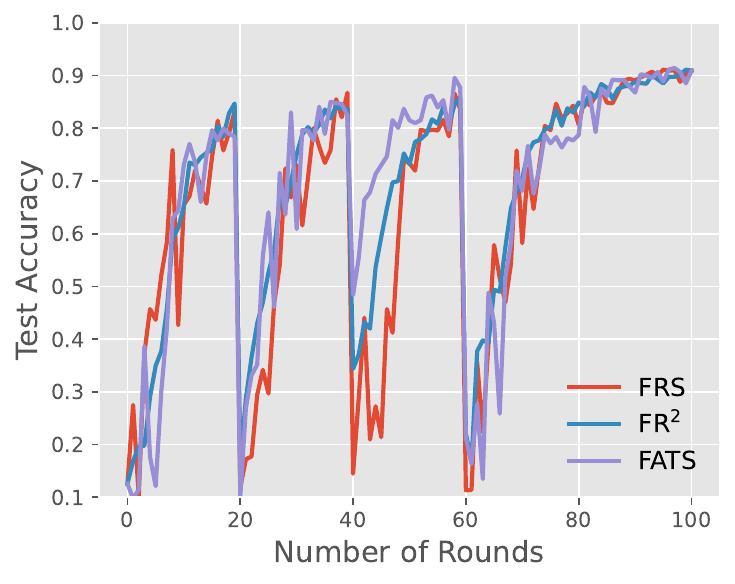}
         \caption{MNIST: Sampling Unlearning}
     \end{subfigure}
     \hfill
     \begin{subfigure}[b]{0.49\linewidth}
        \centering
        \includegraphics[width=\textwidth]{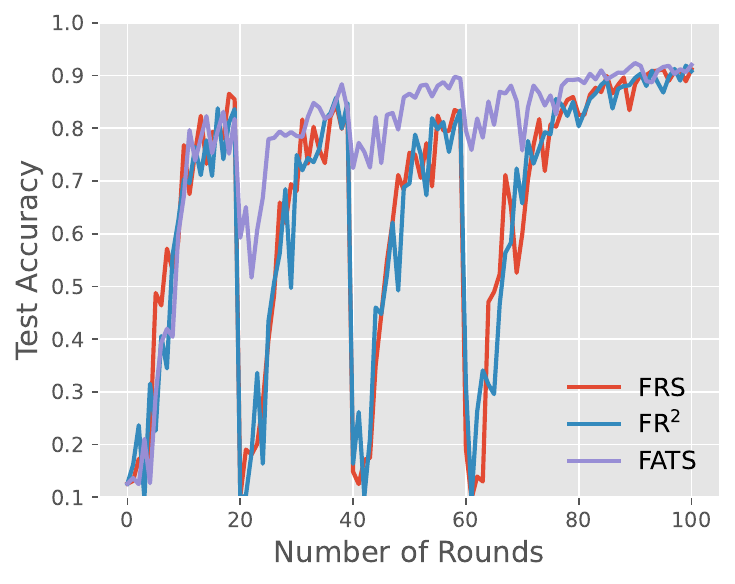}
        \caption{MNIST: Client Unlearning}
     \end{subfigure}
     \vfill
     \begin{subfigure}[b]{0.49\linewidth}
         \centering
         \includegraphics[width=\textwidth]{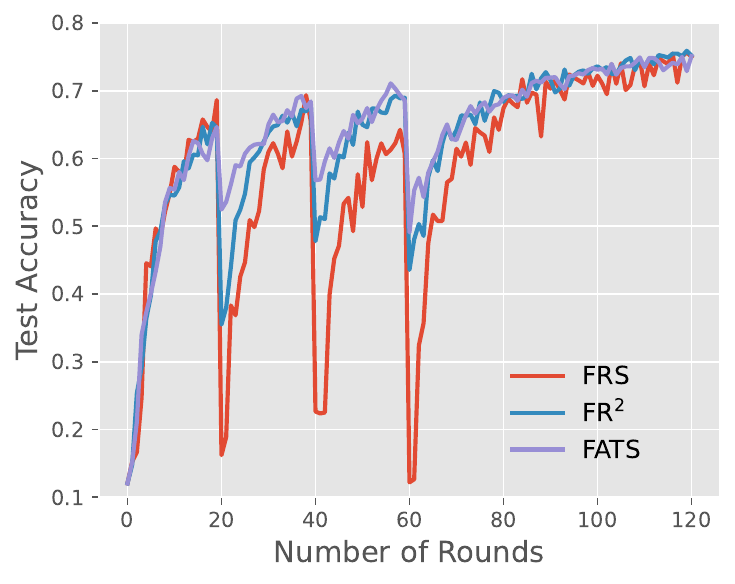}
         \caption{FEMNIST: Sample Unlearning}
     \end{subfigure}
     \hfill
     \begin{subfigure}[b]{0.49\linewidth}
         \centering
         \includegraphics[width=\textwidth]{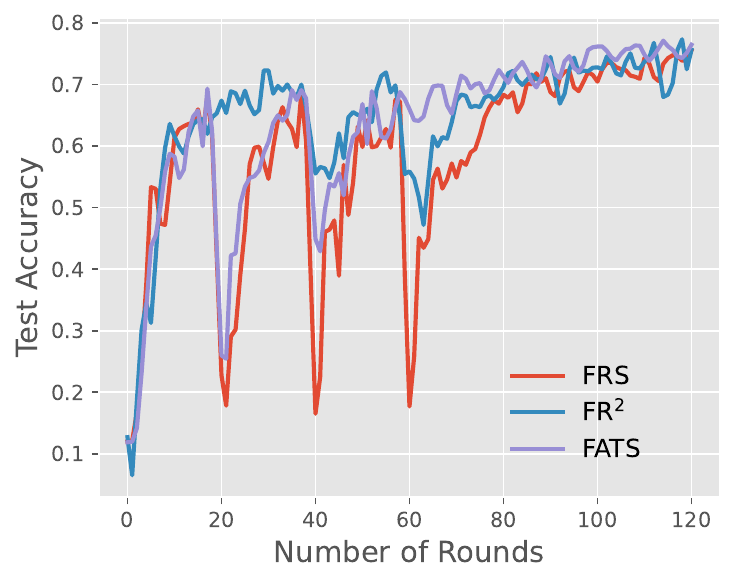}
         \caption{FEMNIST: Client Unlearning}
     \end{subfigure}
     \caption{Comparison of the test accuracy of different methods and their changes under streaming unlearning requests.}
     \label{fig:streaming}
\end{figure*}

\subsection{Additional Results for Unlearning Efficiency}\label{sec:add_efficiency}

We also tested the unlearning efficiency of \texttt{FATS} on the Shakespeare dataset. The results are shown in Figure~\ref{fig:efficiency2} and convey the same observations made in Section~\ref{sec:unlearning_efficiency}.

\subsection{Additional Results for Utility v.s. Efficiency.}\label{sec:add_UvE}

We showcase our additional results for the trade-off between learning utility and unlearning efficiency on MNIST in Figure~\ref{fig:UvsE2}.

\subsection{Performance Evaluation over A Stream of Unlearning Requests}\label{sec:add_streaming}

To provide a more comprehensive performance evaluation, we also investigate the streaming unlearning setting. We evaluate the performance of different methods on MNIST and FEMNIST datasets under streaming unlearning requests, where we sequentially issue multiple requests to remove data samples or clients from the training process. We use the same setting as described in Section~\ref{sec:performance}. Figure~\ref{fig:streaming} shows the results. The results are consistent with the ones presented in Section~\ref{sec:performance}. As we can observe from Figure~\ref{fig:streaming}, our proposed method \texttt{FATS} achieves the highest test accuracy among all methods and maintains a stable performance under different unlearning scenarios. The baseline methods, \texttt{FR}$^2$ and \texttt{FRS}, suffer from significant drops in test accuracy when unlearning requests are issued and thus take more time to recover from the unlearning. This indicates that \texttt{FATS} is more effective and efficient in handling streaming unlearning requests than the other methods.

\section{Omitted Proofs for Unlearning Analysis}

\subsection{Proof of Lemma~\ref{le-stability}}\label{sec:le-stability}

\begin{proof}[Proof of Lemma~\ref{le-stability}]
    We start with the sample-level TV-stability. We denote by $\mu_{(M,K,N,b)}$ the per-round sampling probability measure for $\mathcal{D}$, including both client and mini-batch sampling in each round. 
    Let $\mu^{\otimes R}_{(M,K,N,b)}$ denote the product measure of $R$ copies of $\mu_{(M,K,N,b)}$. Analogously, we define $\mu_{(M,K,N-1,b)}$ and $\mu^{\otimes R}_{(M,K,N-1,b)}$ for $\mathcal{D}^\prime$. We extend the $\sigma$-algebra for probability space with measure $\mu_{(M,K,N-1,b)}$ to obtain measure $\mu^\prime_{(M,K,N,b)}$ such that it has a common probability space with $\mu_{(M,K,N,b)}$. To this end, we define $\mu^\prime_{(M,K,N,b)}$ as follows: for any feasible pair of $(\mathcal{P}^{[r]},\mathcal{B}^{[r]})$, 
    \begin{equation*}
        \begin{aligned}
            \mu^\prime_{(M,K,N,b)}(\mathcal{P}^{[r]},\mathcal{B}^{[r]})=\begin{cases}
                0, \qquad\,\,\text{if $X_{u}\in\mathcal{B}_{k_u}^{[r,i]}$ for some $i\in[E]$,} \\
                \mu_{(M,K,N-1,b)}(\mathcal{P}^{[r]}, \mathcal{B}^{[r]}), \quad\text{otherwise.}
            \end{cases}
        \end{aligned}
    \end{equation*}
    Similarly, we extend the $\sigma$-algebra for the product space with measure $\mu_{(M,K,N-1,b)}^{\otimes R}$ to get $\mu_{(M,K,N,b)}^{\prime\otimes R}$. 
    
    We use $L(\cdot)$ to denote the output model of learning algorithm $\mathcal{L}(\cdot)$. Note that for fixed initial model $\theta^{(0)}$ and other parameters, $L(\mathcal{D})$ and $L(\mathcal{D}^\prime)$ are the same deterministic map from a sampling history $\mathcal{H}\coloneqq[(\mathcal{P}^{[1]}, \mathcal{B}^{[1]}), (\mathcal{P}^{[2]}, \mathcal{B}^{[2]}), \ldots, (\mathcal{P}^{[R]}, \mathcal{B}^{[R]})]$
    to an output model parameter. The discrepancy between $L(\mathcal{D})$ and $L(\mathcal{D}^\prime)$ arises from the distinct measures on the input space that is induced by the sampling histories. Therefore, the total variation distance between $L(\mathcal{D})$ and $L(\mathcal{D}^\prime)$ is equal to the total variation distance between the push-forward measures $L(\mathcal{D})\#\mu_{(M,K,N,b)}^{\otimes R}$ and $L(\mathcal{D}^\prime)\#\mu_{(M,K,N,b)}^{\prime\otimes R}$, which can be bounded above by the total variation distance between $\mu_{(M,K,N,b)}^{\otimes R}$ and $\mu_{(M,K,N,b)}^{\prime\otimes R}$ due to the equivalence of $L(\mathcal{D})$ and $L(\mathcal{D}^\prime)$ and the data processing inequality. Let $\overline{\mathcal{H}}$ be a sampling history that contains at least one pair $(\overline{\mathcal{P}}^{[r]},\overline{\mathcal{B}}^{[r]})$ such that $k_u\in\overline{\mathcal{P}}^{[r]}$ and $X_{k_u}^{(N)}\in\overline{\mathcal{B}}_{k_u}^{[r]}$, which means that the target sample was involved in some round $r$. The total variation distance can be bounded by 
    \begin{flalign}
        &\TV(L(\mathcal{D}), L(\mathcal{D}^\prime))
        \le\TV(\mu_{(M,K,N,b)}^{\otimes R}, \mu_{(M,K,N,b)}^{\prime\otimes R})\notag\\
        &=\sup_{\text{$\mathcal{H}$ over all choices}}|\mu_{(M,K,N,b)}^{\otimes R}(\mathcal{H})-\mu_{(M,K,N,b)}^{\prime\otimes R}(\mathcal{H})|\notag\\
        &=\mu_{(M,K,N,b)}^{\otimes R}(\overline{\mathcal{H}})\le R\cdot\mu_{(M,K,N,b)}(\overline{\mathcal{P}}^{[r]}, \overline{\mathcal{B}}^{[r]})\notag\\
        &\le R\cdot(\frac{K}{M}\cdot E\cdot\frac{b}{N})\le\frac{TKb}{MN},\notag
    \end{flalign}
    where the first inequality is due to the data-processing inequality and the last inequality is due to $R\le\frac{T}{E}$. By the definition of TV distance, we have $\TV(L(\mathcal{D}),L(\mathcal{D}^\prime))\le 1$ as a trivial bound. Noting that $K=\frac{\rho_C\cdot E\cdot M}{T}$ and $b=\frac{\rho_S\cdot N}{\rho_C\cdot E}$, we obtain the result that the output of \texttt{FATS} is $\min\{\rho_S, 1\}$ sample-level TV-stable. 

    Next, we move on to the client-level TV-stability. Unlike the sample-level case where a re-computation is triggered only when the target sample was involved, in client-level unlearning, a re-computation would be triggered if the target client was involved. Thus, for client-level unlearning, we focus on the client sampling history, i.e., $\mathcal{H}_{c}=\{\mathcal{P}^{[1]}, \mathcal{P}^{[2]}, \ldots, \mathcal{P}^{[R]}\}$.
    Let $\mathcal{C}$ be the original set of all clients and $\mathcal{C}^\prime\coloneqq\mathcal{C}\setminus\mathbb\{k_u\}$. We denote by $\nu_{(M,K)}$ the per-round sampling probability measure for $\mathcal{C}$. Then $\nu_{(M,K)}^{\otimes R}$ denotes the product measure of $R$ copies of $\nu_{(M,K)}$. We similarly define $\nu_{(M-1,K)}$ and $\nu_{(M-1,K)}^{\otimes R}$ for $\mathcal{C}^\prime$. We extend the $\sigma$-algebra for the probability space with measure $\nu_{(M-1,K)}$ to get $\nu^\prime_{(M,K)}$ so that it has a common probability space with $\nu_{(M,K)}$. We define $\nu^\prime_{(M,K)}$ as follows: for any possible client multiset $\mathcal{P}^{[r]}$,
    \begin{equation*}
        \begin{aligned}
            \nu^\prime_{(M,K)}(\mathcal{P}^{[r]})=
            \begin{cases}
                0&, \text{if $k_u\in\mathcal{P}^{[r]}$} \\
                \nu_{(M,K)}(\mathcal{P}^{[r]})&, \text{otherwise}
            \end{cases}.
        \end{aligned}
    \end{equation*}
    We similarly extend the $\sigma$-algebra for the product space with measure $\nu_{(M-1,K)}^{\otimes R}$ to get $\nu_{(M,K)}^{\prime\otimes R}$. Let $\overline{\mathcal{H}_c}$ be a client sampling history that contains at least one $\overline{\mathcal{P}}^{[r]}$ for $r\in[R]$ such that $k_u\in\overline{\mathcal{P}}^{[r]}$. Following a similar argument as in the sample-level case and using the data processing inequality, we can bound the total variation distance between $L(\mathcal{D}(\mathcal{C}))$ and $L(\mathcal{D}(\mathcal{C}^\prime))$ by 
    \begin{flalign}
        &\TV(L(\mathcal{D}(\mathcal{C})), L(\mathcal{D}(\mathcal{C}^\prime)))
        \le\TV(\nu_{(M,K)}^{\otimes R}, \nu_{(M,K)}^{\prime\otimes R})\notag\\
        &=\sup_{\text{$\mathcal{H}_{c}$ over all choices}}|\nu_{(M,K)}^{\otimes R}(\mathcal{H}_{c})-\nu_{(M,K)}^{\prime\otimes R}(\mathcal{H}_{c})|\notag\\
        &=\nu_{(M,K)}^{\otimes R}(\overline{\mathcal{H}_c})\le R\cdot\nu_{(M,K)}(\overline{\mathcal{P}}^{[r]})\le R\cdot\frac{K}{M}\le\frac{TK}{EM},\notag
    \end{flalign}
    where the first inequality is due to the data-processing inequality and the last inequality is due to $R\le\frac{T}{E}$. By definition of TV distance, we have $\TV(L(\mathcal{D}(\mathcal{C})), L(\mathcal{D}(\mathcal{C}^\prime)))\le 1$ as a trivial bound. Noticing that $K=\frac{\rho_C\cdot E\cdot M}{T}$, we obtain the result that the output of \texttt{FATS} is $\min\{\rho_C, 1\}$ client-level TV-stable. 
 \end{proof}

\subsection{Proof of Theorem~\ref{th-unlearning}}\label{sec:th-unlearning}

\begin{proof}[Proof of Theorem~\ref{th-unlearning}]
    We adopt the same notation as in the proof of Lemma~\ref{le-stability} and we first consider the one deletion case. For sample-level unlearning, we essentially substitute some mini-batches drawn by client $k_u$ in each round that $k_u$ participated in. To simplify the presentation, we use $\texttt{SU}_i:(\mathcal{D}_{k_u})^{b} \to (\mathcal{D}_{k_u}\setminus\{X_u\})^{b}$ to denote one iteration of $\texttt{FATS-SU}$ for handling a mini-batch drawn by $k_u$, and $\texttt{SU}_r:(\mathcal{D}_{k_u})^{b\times E} \to (\mathcal{D}_{k_u}\setminus\{X_u\})^{b\times E}$ to denote all such operations in one communication round. For any non-empty $\mathcal{B}_{k_u}^{[r]}\in(\mathcal{D}_{k_u})^{b\times E}$ generated by $\mu_{(M,K,N,b)}$, we observe that the mapping $\texttt{SU}_r$ acts \textit{component-wise} and \textit{symmetrically}, which means that
    \[
    \texttt{SU}_r(\mathcal{B}_{k_u}^{[r]})=\left(\texttt{SU}_i(\mathcal{B}_{k_u}^{[r,1]}),\texttt{SU}_i(\mathcal{B}_{k_u}^{[r,2]}),\ldots,\texttt{SU}_i(\mathcal{B}_{k_u}^{[r,E]})\right).
    \]
    Define $\mu_{(M,K,N,b)}^{\texttt{SU}_r}\coloneqq\texttt{SU}_r\#\mu_{(M,K,N,b)}$. We first show in Claim~\ref{cl-transport} that $\texttt{SU}_r$ transports $\mu_{(M,K,N,b)}$ to $\mu_{(M,K,N-1,b)}$, i.e., $\mu_{(M,K,N,b)}^{\texttt{SU}_r}=\mu_{(M,K,N-1,b)}$. The proof of Claim~\ref{cl-transport} is in Appendix~\ref{sec:cl-transprt}.
    \begin{claim}\label{cl-transport}
        For any possible pair $(\mathcal{P}^{[r]},\mathcal{B}^{[r]})$, it holds that 
        \[
            \mu_{(M,K,N,b)}^{\texttt{SU}_r}(\mathcal{P}^{[r]},\mathcal{B}^{[r]})=\mu_{(M,K,N-1,b)}(\mathcal{P}^{[r]},\mathcal{B}^{[r]}).
        \]
    \end{claim}
    
 We proceed to describe the coupling constructed by the unlearning operations. Suppose we first get a sampling history 
 \[
    \mathcal{H}=((\mathcal{P}^{[1]}, \mathcal{B}^{[1]}), (\mathcal{P}^{[2]}, \mathcal{B}^{[2]}), \ldots, (\mathcal{P}^{[R]}, \mathcal{B}^{[R]}))
 \]
 according to the measure $\mu_{(M,K,N,b)}^{\otimes R}$. Furthermore, we consider another sampling history $\mathcal{H}^\prime$ whose each element, denoted as $(\mathcal{P}^{\prime[r]},\mathcal{B}^{\prime[r]})$, is defined as follows. If $X_{u}$ is not contained in any mini-batch in $\mathcal{B}_{k_u}^{[r]}$, then $(\mathcal{P}^{\prime[r]},\mathcal{B}^{\prime[r]})=(\mathcal{P}^{[r]},\mathcal{B}^{[r]})$. Otherwise, $(\mathcal{P}^{\prime[r]},\mathcal{B}^{\prime[r]})$ is sampled from $\mu_{(M,K,N-1,b)}$. This way, we obtain a pair of coupled sampling histories $\mathcal{H}$ and $\mathcal{H}^\prime$. Following directly from Claim~\ref{cl-transport}, we have the following claim for the coupled history. 
 \begin{claim}\label{cl-coupling}
     For any possible sampling history $H$, it holds that $\mathbb{P}(\mathcal{H}=H)=\mu_{(M,K,N,b)}^{\otimes R}$ and $\mathbb{P}(\mathcal{H}^\prime=H)=\mu_{(M,K,N-1,b)}^{\otimes R}$.
 \end{claim}
 Claim~\ref{cl-coupling} states that our unlearning operations in \texttt{FATS-SU} indeed transport the per-round sample and client sampling probability measure $\mu_{(M,K,N,b)}^{\otimes R}$ for the original dataset $\mathcal{D}$ to the measure $\mu_{(M,K,N-1,b)}^{\otimes R}$ for the updated dataset $\mathcal{D}^\prime$, which guarantees sample-level exact unlearning of \texttt{FATS-SU}.
 Next, we show in Claim~\ref{cl-re-compute} that the probability of disagreement under the above coupling, i.e., the probability that a re-computation occurs, is bounded by sample-level TV-stability $\rho_S$. The proof of Claim~\ref{cl-re-compute} is in Appendix~\ref{sec:cl-re-compute}.
 \begin{claim}\label{cl-re-compute}
     For the $\rho_S$ sample-level TV-stable \texttt{FATS}, under the coupling of $(\mathcal{H},\mathcal{H}^\prime)$, we have $\mathbb{P}_{(\mathcal{H},\mathcal{H}^\prime)}(\mathcal{H}\neq\mathcal{H}^\prime)\le\rho_S$.
 \end{claim}
 Finally, from Remark~\ref{rem-TV}, for $w$ sample unlearning requests, the re-compute probability is at most $w\cdot\rho_S$.

  The proof for the client-level unlearning case follows a similar argument as sample-level unlearning. The main difference is that we only focus on client sampling, just as we did for Lemma~\ref{le-stability}. 
\end{proof}

\subsection{Proof of Claim~\ref{cl-transport}}\label{sec:cl-transprt}

\begin{proof}[Proof of Claim~\ref{cl-transport}]
    We concentrate on the mini-batch sampling measure at client $k_u$, since the unlearning operation only affects the mini-batches sampled by client $k_u$. Specifically, we denote by $\xi_{(N,b)}$ the probability measure of $k_u$'s sampling a mini-batch of size $b$ from $\mathcal{D}_{k_u}$ and by $\xi^{\texttt{dul}}_{(N,b)}\coloneqq\texttt{dul}\#\xi_{(N,b)}$. Similarly, we denote by $\xi_{(N-1,b)}$ the probability measure for sampling a $b$ size mini-batch from $\mathcal{D}_{k_u}\setminus\{X_{u}\}$. Let $B\in(\mathcal{D}_{k_u})^{b}$ be any possible mini-batch. We show $\xi^{\texttt{dul}}_{(N,b)}(B)=\xi_{(N-1,b)}(B)$ for each local iteration by two cases.
    \paragraph*{Case 1} If the verification fails, i.e., a re-computation is triggered, the algorithm will re-sample $B\sim\xi_{(N-1,b)}$. Thus, $\xi^{\texttt{dul}}_{(N,b)}(B)=\xi_{(N-1,b)}(B)$ holds trivially for this case.
    \paragraph*{Case 2} If the verification succeeds. Then the deleted data $X_u$ is not contained in $B$. The measure $\xi^{\texttt{dul}}_{(N,b)}$ is therefore just the probability under the original sampling measure $\xi_{(N,b)}$ conditioned on the event that $X_u\notin B$. Therefore,
    \begin{flalign}
        \xi^{\texttt{dul}}_{(N,b)}(B)=\xi_{(N,b)}(B|X_u\notin B)=\frac{\xi_{(N,b)}(B\cap\{X_u\notin B\})}{\xi_{(N,b)}(\{X_u\notin B\})},\notag
    \end{flalign}
    where, by direct computation, $\xi_{(N,b)}(\{X_u\notin B\})=1-\xi_{(N,b)}(\{X_u\in B\})=1-\frac{\binom{N-1}{b-1}}{\binom{N}{b}}=1-\frac{b}{N}$. To calculate the numerator, we consider two possible choices of $B$. First, if $X_u\in B$, then $\xi_{(N,b)}(B\cap\{X_u\notin B\})=0$, which gives us that $\xi^{\texttt{dul}}_{(N,b)}(B)=0=\xi_{(N-1,b)}(B)$. For the choice that $X_u\notin B$,
    \begin{flalign}
        \xi^{\texttt{dul}}_{(N,b)}(B)
        &=\frac{\xi_{(N-1,b)}(B)}{\xi_{(N-1,b)}(\{X_u\notin B\})}=\frac{1/\binom{N}{b}}{1-\frac{b}{N}}\notag\\
        &=\frac{N}{N-b}\frac{(N-b)!b!}{N!}=\frac{(N-b-1)!b!}{(N-1)!}\notag\\
        &=\frac{1}{\binom{N-1}{b}}=\xi_{(N-1,b)}(B).\notag
    \end{flalign}
    Therefore, $\xi^{\texttt{dul}}_{(N,b)}(B)=\xi_{(N-1,b)}(B)$ also holds for Case~2.
    \end{proof}

\subsection{Proof of Claim~\ref{cl-re-compute}}\label{sec:cl-re-compute}

\begin{proof}[Proof of Claim~\ref{cl-re-compute}]
 \begin{flalign}
    &\mathbb{P}_{(\mathcal{H},\mathcal{H}^\prime)}(\mathcal{H}\neq\mathcal{H}^\prime)\notag\\
    &=\mathbb{P}_{(\mathcal{H},\mathcal{H}^\prime)}(\exists r\in[R] \text{ s.t. }(\mathcal{P}[r],\mathcal{B}[r])\neq(\mathcal{P}^{\prime[r]},\mathcal{B}^{\prime[r]}))\notag\\
    &=\mathbb{P}_{\mathcal{H}}(\exists r\in[R], i\in[E] \text{ s.t. }X_{u}\in\mathcal{B}_{k_u}^{[r,i]}) \le\frac{TKb}{MN}=\rho_S.\notag
 \end{flalign}
 \end{proof}

\section{Omitted Proofs for Convergence Analysis}
\subsection{Proof of Lemma~\ref{le-convergence}}\label{sec:le-convergence}

\begin{proof}[Proof of Lemma~\ref{le-convergence}]
    Our algorithm involves two sources of stochasticity. The first one arises from the stochastic gradient, while the second one stems from the client sampling. To differentiate them, we adopt the notation $\mathbb{E}_{\mathcal{B}^{(t)}}\coloneqq\mathbb{E}_{\mathcal{B}^{(t)}_1,\ldots,\mathcal{B}^{(t)}_M}$ to indicate the expectation with respect to the randomness of mini-batch sampling in iteration $t$, and use $\mathbb{E}_{\mathcal{P}^{(t)}}$ to eliminate the randomness of client sampling in iteration $t$. Let $\mathbb{E}_t$ be the expectation with respect to all of the randomness until time $t$, then $\mathbb{E}_t=\mathbb{E}_{\mathcal{B}^{(t)}}\mathbb{E}_{\mathcal{P}^{(t)}}\mathbb{E}_{t-1}$. We drop the time indicator $t$ and simply use $\mathbb{E}$ to denote the expectation with respect to all the relevant randomness when it is clear from the context.
    
    Using the update rule $\theta_k^{(t)}\gets\theta_k^{(t-1)}-\eta\cdot \tilde{g}_k^{(t)}$ and the $L$-smoothness of the loss function $f$ (Assumption~\ref{asp-smoothness}), we derive the following inequality:
\begin{equation}
    F(\theta^{(t)})-F(\theta^{(t-1)})\le-\eta\langle \nabla F(\theta^{(t-1)}), \tilde{g}^{(t)}\rangle+\frac{\eta^2L}{2}\|\tilde{g}^{(t)}\|_2^2.\notag
\end{equation}
Taking expectation on both sides, we obtain
\begin{flalign}
    \mathbb{E}[F(\theta^{(t)})-F(\theta^{(t-1)})]\le-\eta\mathbb{E}[\langle \nabla F(\theta^{(t-1)}), \tilde{g}^{(t)}\rangle]+\frac{\eta^2L}{2}\mathbb{E}[\|\tilde{g}^{(t)}\|_2^2].\notag
\end{flalign}
Averaging over all iterations $t=1,2,\ldots, T$, we get
\begin{flalign}\label{eq-lossupdate}
    \frac{1}{T}\sum_{t=1}^{T}\mathbb{E}[F(\theta^{(t)})-F(\theta^{(t-1)})]\le-\frac{\eta}{T}\sum_{t=1}^{T}\mathbb{E}[\langle \nabla F(\theta^{(t-1)}), \tilde{g}^{(t)}\rangle]+\frac{\eta^2L}{2T}\sum_{t=1}^{T}\mathbb{E}[\|\tilde{g}^{(t)}\|_2^2].
\end{flalign}
To further bound the terms on the right hand side of the above inequality, we need the following results.
\begin{claim}\label{cl-innerprod}
 The expected inner product between the stochastic and the full-batch gradients in iteration $t$ can be bounded as follows:
    \begin{flalign}
        -\eta\mathbb{E}[\langle \nabla F(\theta^{(t-1)}), \tilde{g}^{(t)}\rangle]\le-\frac{\eta}{2}\mathbb{E}[\|\nabla F(\theta^{(t-1)})\|_2^2]-\frac{\eta}{2}\mathbb{E}[\|\frac{1}{M}\sum_{k=1}^Mg_k^{(t)}\|_2^2]+\frac{\eta L^2}{2M}\sum_{k=1}^M\mathbb{E}[\|\theta^{(t-1)}-\theta_k^{(t-1)}\|_2^2].\notag
    \end{flalign}
\end{claim}

\begin{claim}\label{cl-variance}
The averaged distance of local models from their virtual average during the learning process holds for
\begin{flalign}
    \frac{1}{TM}\sum_{t=1}^{T}\sum_{k=1}^M\mathbb{E}[\|\theta^{(t-1)}-\theta_k^{(t-1)}\|_2^2]\le \frac{E(K+1)\eta^2G^2}{Kb}+\frac{2\eta^2\lambda E(E-1)}{T}\sum_{t=1}^{T}\mathbb{E}[\|\frac{1}{M}\sum_{k=1}^M g_k^{(t)}\|_2^2]\notag
\end{flalign}
\end{claim}

\begin{claim}\label{cl-stgradnorm}
    The expected squared norm of the averaged stochastic gradient of iteration $t$ can be bounded as follows:
    \begin{equation*}
        \mathbb{E}[\|\tilde{g}^{(t)}\|_2^2]\le\frac{G^2}{Kb}+\lambda\mathbb{E}[\|\frac{1}{M}\sum_{k=1}^{M}g_k^{(t)}\|_2^2]
    \end{equation*}
\end{claim}

We continue the proof by utilizing Claim~\ref{cl-innerprod}, claim~\ref{cl-variance} and claim~\ref{cl-stgradnorm} to further bound \eqref{eq-lossupdate} as follows:
\begin{flalign*}
    &\frac{1}{T}\sum_{t=1}^{T}\mathbb{E}[F(\theta^{(t)})-F(\theta^{(t-1)})]\notag\\
    &\le\frac{1}{T}\sum_{t=1}^{T}(-\eta\mathbb{E}[\langle \nabla F(\theta^{(t-1)}), \tilde{g}^{(t)}\rangle])+\frac{1}{T}\sum_{t=1}^{T}\frac{\eta^2L}{2}\mathbb{E}[\|\tilde{g}^{(t)}\|_2^2]\notag\\
    &\le\frac{1}{T}\sum_{t=1}^{T}\Big(-\frac{\eta}{2}\mathbb{E}[\|\nabla F(\theta^{(t-1)})\|_2^2]-\frac{\eta}{2}\mathbb{E}[\|\frac{1}{M}\sum_{k=1}^Mg_k^{(t)}\|_2^2]+\frac{\eta L^2}{2M}\sum_{k=1}^M\mathbb{E}[\|\theta^{(t-1)}-\theta_k^{(t-1)}\|_2^2]\Big)+\frac{1}{T}\sum_{t=1}^{T}(\frac{\eta^2L}{2}(\frac{G^2}{Kb}+\lambda\mathbb{E}[\|\frac{1}{M}\sum_{k=1}^{M}g_k^{(t)}\|_2^2]))\\
    &=-\frac{\eta}{2T}\sum_{t=1}^{T}\mathbb{E}[\|\nabla F(\theta^{(t-1)})\|_2^2]+(-\frac{\eta}{2}+\eta^3L^2\lambda E(E-1)+\frac{\eta^2\lambda L}{2})\frac{1}{T}\sum_{t=1}^{T}\mathbb{E}[\|\frac{1}{M}\sum_{k=1}^M g_k^{(t)}\|_2^2]+ \frac{\eta^3L^2G^2E(K+1)}{2Kb}+\frac{\eta^2LG^2}{2Kb}\\
    &\le -\frac{\eta}{2T}\sum_{t=1}^{T}\mathbb{E}[\|\nabla F(\theta^{(t-1)})\|_2^2]+ \frac{\eta^3L^2G^2E(K+1)}{2Kb}+\frac{\eta^2LG^2}{2Kb},
\end{flalign*}
where the last inequality is due to the condition that $-\frac{\eta}{2}+\eta^3L^2\lambda E(E-1)+\frac{\eta^2\lambda L}{2}<0$. 

By rearranging, we get 
\begin{flalign*}
    \frac{1}{T}\sum_{t=1}^{T}\mathbb{E}[\|\nabla F(\theta^{(t-1)})\|_2^2]\le\frac{2(F(\theta^{(0)})-F^*)}{\eta T}+\frac{\eta^2L^2G^2E(K+1)}{Kb}+\frac{\eta LG^2}{Kb}.
\end{flalign*}
Finally, by noting that $K=\frac{\rho_CEM}{T}$ and $b=\frac{\rho_SN}{\rho_CE}$, we can conclude the proof.
\end{proof}

\subsection{Proof of Claim~\ref{cl-innerprod}}\label{sec:cl-innerprod}
\begin{proof}[Proof of Claim~\ref{cl-innerprod}]
    \begin{flalign*}
        &\quad-\eta\mathbb{E}_{t}[\langle \nabla F(\theta^{(t-1)}), \tilde{g}^{(t)}\rangle]\\
        &=-\eta\mathbb{E}_{t-1}\mathbb{E}_{\mathcal{B}^{(t)}}[\mathbb{E}_{\mathcal{P}^{(t)}}[\langle \nabla F(\theta^{(t-1)}), \tilde{g}^{(t)}\rangle]]\\
        &=-\eta\mathbb{E}_{t-1}\mathbb{E}_{\mathcal{B}^{(t)}}[\mathbb{E}_{\mathcal{P}^{(t)}}[\langle \nabla F(\theta^{(t-1)}),\frac{1}{K}\sum_{k\in\mathcal{P}^{(t)}}\tilde{g}_k^{(t)}\rangle]]\\
        &\overset{\text{\cone}}{=}-\eta\mathbb{E}_{t-1}\mathbb{E}_{\mathcal{P}^{(t)}}[\mathbb{E}_{\mathcal{B}^{(t)}}[\langle \nabla F(\theta^{(t-1)}),\frac{1}{K}\sum_{k\in\mathcal{P}^{(t)}}\tilde{g}_k^{(t)}\rangle]]\\
        &=-\eta\mathbb{E}_{t-1}[\langle \nabla F(\theta^{(t-1)}),\mathbb{E}_{\mathcal{P}^{(t)}}[\frac{1}{K}\sum_{k\in\mathcal{P}^{(t)}}\mathbb{E}_{\mathcal{B}^{(t)}}[\tilde{g}_k^{(t)}]]\rangle]\\
        &=-\eta\mathbb{E}_{t-1}[\langle \nabla F(\theta^{(t-1)}),\frac{1}{K}\mathbb{E}_{\mathcal{P}^{(t)}}[\sum_{k\in\mathcal{P}^{(t)}}g_k^{(t)}]\rangle]\\
        &=-\eta\mathbb{E}_{t-1}[\langle \nabla F(\theta^{(t-1)}),\frac{1}{K}[K\sum_{k=1}^M\frac{1}{M}g_k^{(t)}]\rangle]\\
        &\overset{\text{\ctwo}}{=}-\frac{\eta}{2}\mathbb{E}_{t-1}[\|\nabla F(\theta^{(t-1)})\|_2^2]-\frac{\eta}{2}\mathbb{E}_{t-1}[\|\frac{1}{M}\sum_{k=1}^M g_k^{(t)}\|_2^2]+\frac{\eta}{2}\mathbb{E}_{t-1}[\|\nabla F(\theta^{(t-1)})-\frac{1}{M}\sum_{k=1}^M g_k^{(t)}\|_2^2]\\
        &=-\frac{\eta}{2}\mathbb{E}_{t-1}\|\nabla F(\theta^{(t-1)})\|_2^2]-\frac{\eta}{2}\mathbb{E}_{t-1}[\|\frac{1}{M}\sum_{k=1}^M g_k^{(t)}\|_2^2]+\frac{\eta}{2}\mathbb{E}_{t-1}[\|\frac{1}{M}\sum_{k=1}^M (\nabla F_k(\theta^{(t-1)})-g_k^{(t)})\|_2^2]\\
        &\overset{\text{\cthree}}{\le}-\frac{\eta}{2}\mathbb{E}_{t-1}[\|\nabla F(\theta^{(t-1)})\|_2^2]-\frac{\eta}{2}\mathbb{E}_{t-1}[\|\frac{1}{M}\sum_{k=1}^M g_k^{(t)}\|_2^2]+\frac{\eta}{2}\frac{1}{M}\sum_{k=1}^M\mathbb{E}_{t-1}[\|\nabla F_k(\theta^{(t-1)})-\nabla F_k(\theta_k^{(t-1)})\|_2^2]\\
        &\overset{\text{\cfour}}{\le}-\frac{\eta}{2}\mathbb{E}_{t-1}[\|\nabla F(\theta^{(t-1)})\|_2^2]-\frac{\eta}{2}\mathbb{E}_{t-1}[\|\frac{1}{M}\sum_{k=1}^M g_k^{(t)}\|_2^2]+\frac{\eta}{2}\frac{L^2}{M}\sum_{k=1}^M\mathbb{E}_{t-1}[\|\theta^{(t-1)}-\theta_k^{(t-1)}\|_2^2],
    \end{flalign*}
    where \text{\cone} is due to the fact that the randomness of $\mathcal{B}^{(t)}$ and $\mathcal{P}^{(t)}$ are independent, \text{\ctwo} is due to the equation $2\langle a,b\rangle=\|a\|_2^2+\|b\|_2^2-\|a-b\|_2^2$, \text{\cthree} holds because of convexity of $\|\cdot\|_2$, and \text{\cfour} follows from $L$-smoothness of loss function $f$ (Assumption~\ref{asp-smoothness}).
\end{proof}

\subsection{Proof of Claim~\ref{cl-variance}}\label{sec:cl-variance}

\begin{proof}[Proof of Claim~\ref{cl-variance}]
   Define $t_c\coloneqq\lfloor \frac{t-1}{E}\rfloor E$, which denotes the last iteration before entering the current round of iteration $t$. The local model at client $k$ in any iteration $t>t_c$ can be expressed as
    \begin{flalign}\label{eq-updaterule}
        \theta_k^{(t)}=\theta_k^{(t-1)}-\eta\tilde{g}_k^{(t)}=\theta_k^{(t-2)}-[\eta\tilde{g}_k^{(t-2)}+\eta\tilde{g}_k^{t-1}]=\ldots=\theta_k^{t_c}-\sum_{\tau=t_c+1}^{t}\eta\tilde{g}_k^{(\tau)}.
    \end{flalign}
    From \eqref{eq-updaterule}, we compute the virtual average model in $t$-th iteration as follows:
    \begin{equation}
        \theta^{(t)}=\theta^{(t_c)}-\frac{\eta}{K}\sum_{k\in\mathcal{P}^{(t)}}\sum_{\tau=t_c+1}^{t}\tilde{g}_k^{(\tau)}.\notag
    \end{equation}
    We express $t$ as $t=sE+r$, where $s\in\{0,1,\ldots,\lfloor\frac{T-1}{E}\rfloor\}$ denotes the indices of communication round and $r\in[E]$ denotes the indices of local updates. Observe that for $t_c< t\le t_c+E$, $\mathbb{E}[\|\theta^{(t-1)}-\theta_k^{(t-1)}\|_2^2$ does not depend on time steps $t\le t_c$ for all client $k\in[M]$. Thus, we obtain the following for all iterations $1\le t\le T$:
    \begin{flalign*}
        \frac{1}{TM}\sum_{t=1}^{T}\sum_{k=1}^M\mathbb{E}[\|\theta^{(t-1)}-\theta_k^{(t-1)}\|_2^2]=\frac{1}{TM}\sum_{s=0}^{\lfloor\frac{T-1}{E}\rfloor}\sum_{r=1}^{E}\sum_{k=1}^M\mathbb{E}[\|\theta^{(sE+r-1)}-\theta_k^{(sE+r-1)}\|_2^2].
    \end{flalign*}
    We first bound the term $\mathbb{E}[\|\theta^{(sE+r-1)}-\theta_k^{(sE+r-1)}\|_2^2]$ for any fixed tuple of $(s, r, k)$.
    \begin{flalign*}
        &\quad\mathbb{E}[\|\theta^{(sE+r-1)}-\theta_k^{(sE+r-1)}\|_2^2]\\
        &=\mathbb{E}[\|\theta^{(sE)}-\frac{1}{K}\sum_{l\in\mathcal{P}^{(t)}}\sum_{i=1}^{r-1}\eta\tilde{g}_l^{(sE+i)}-\theta^{(sE)}+\sum_{i=1}^{r-1}\eta\tilde{g}_k^{(sE+i)}\|_2^2]\notag\\
        &=\mathbb{E}[\|\sum_{i=1}^{r-1}\eta\tilde{g}_k^{(sE+i)}-\frac{1}{K}\sum_{l\in\mathcal{P}^{(t)}}\sum_{i=1}^{r-1}\eta\tilde{g}_l^{(sE+i)}\|_2^2]\\
        &\overset{\text{\cone}}{\le}2\mathbb{E}[\|\sum_{i=1}^{r-1}\eta\tilde{g}_k^{(sE+i)}\|_2^2+\|\frac{1}{K}\sum_{l\in\mathcal{P}^{(t)}}\sum_{i=1}^{r-1}\eta\tilde{g}_l^{(sE+i)}\|_2^2]\\
        &\overset{\text{\ctwo}}{=}2\mathbb{E}[\|\sum_{i=1}^{r-1}\eta\tilde{g}_k^{(sE+i)}-\mathbb{E}[\sum_{i=1}^{r-1}\eta\tilde{g}_k^{(sE+i)}]\|_2^2+\|\mathbb{E}[\sum_{i=1}^{r-1}\eta\tilde{g}_k^{(sE+i)}]\|_2^2+\|\frac{1}{K}\sum_{l\in\mathcal{P}^{(t)}}\sum_{i=1}^{r-1}\eta\tilde{g}_l^{(sE+i)}-\mathbb{E}[\frac{1}{K}\sum_{l\in\mathcal{P}^{(t)}}\sum_{i=1}^{r-1}\eta\tilde{g}_l^{(sE+i)}]\|_2^2+\|\mathbb{E}[\frac{1}{K}\sum_{l\in\mathcal{P}^{(t)}}\sum_{i=1}^{r-1}\eta\tilde{g}_l^{(sE+i)}]\|_2^2]\\
        &=2\mathbb{E}\bigg[\|\sum_{i=1}^{r-1}\eta(\tilde{g}_k^{(sE+i)}-g_k^{(sE+i)})\|_2^2+\|\sum_{i=1}^{r-1}\eta g_k^{(sE+i)}\|_2^2+\|\frac{1}{K}\sum_{l\in\mathcal{P}^{(t)}}\sum_{i=1}^{r-1}\eta(\tilde{g}_l^{(sE+i)}-g_l^{(sE+i)})\|_2^2+\|\frac{1}{K}\sum_{l\in\mathcal{P}^{(t)}}\sum_{i=1}^{r-1}\eta g_l^{(sE+i)}\|_2^2\bigg]\\
        &=2\mathbb{E}\Bigg[\sum_{i=1}^{r-1}\eta^2\|\tilde{g}_k^{(sE+i)}-g_k^{(sE+i)}\|_2^2+\sum_{u\neq v}\langle \eta\tilde{g}_k^{(u)}-\eta g_k^{(u)}, \eta\tilde{g}_k^{(v)}-\eta g_k^{(v)}\rangle+\|\sum_{i=1}^{r-1}\eta g_k^{(sE+i)}\|_2^2\notag\\
        &\quad+\frac{1}{K^2}\sum_{l\in\mathcal{P}^{(t)}}\sum_{i=1}^{r-1}\eta^2\|\tilde{g}_l^{(sE+i)}-g_l^{(sE+i)}\|_2^2+\frac{1}{K^2}\sum_{u\neq v, y\neq z}\langle \eta\tilde{g}_y^{(u)}-\eta g_y^{(u)}, \eta\tilde{g}_z^{(v)}-\eta g_z^{(v)}\rangle+\|\frac{1}{K}\sum_{l\in\mathcal{P}^{(t)}}\sum_{i=1}^{r-1}\eta g_l^{(sE+i)}\|_2^2\Bigg]\\
        &\overset{\text{\cthree}}{=}2\mathbb{E}\Bigg[\sum_{i=1}^{r-1}\eta^2\|\tilde{g}_k^{(sE+i)}-g_k^{(sE+i)}\|_2^2+\|\sum_{i=1}^{r-1}\eta g_k^{(sE+i)}\|_2^2+\frac{1}{K^2}\sum_{l\in\mathcal{P}^{(t)}}\sum_{i=1}^{r-1}\eta^2\|\tilde{g}_l^{(sE+i)}-g_l^{(sE+i)}\|_2^2+\|\frac{1}{K}\sum_{l\in\mathcal{P}^{(t)}}\sum_{i=1}^{r-1}\eta g_l^{(sE+i)}\|_2^2\Bigg]\\
        &\overset{\text{\cfour}}{\le} 2\mathbb{E}[\sum_{i=1}^{r-1}\eta^2\|\tilde{g}_k^{(sE+i)}-g_k^{(sE+i)}\|_2^2+(r-1)\eta^2\sum_{i=1}^{r-1}\| g_k^{(sE+i)}\|_2^2+\frac{1}{K^2}\sum_{l\in\mathcal{P}^{(t)}}\sum_{i=1}^{r-1}\eta^2\|\tilde{g}_l^{(sE+i)}-g_l^{(sE+i)}\|_2^2+\frac{(r-1)\eta^2}{K}\sum_{l\in\mathcal{P}^{(t)}}\sum_{i=1}^{r-1}\| g_l^{(sE+i)}\|_2^2]\\
        &=2\sum_{i=1}^{r-1}\eta^2\mathbb{E}[\|\tilde{g}_k^{(sE+i)}-g_k^{(sE+i)}\|_2^2]+2(r-1)\eta^2\sum_{i=1}^{r-1}\mathbb{E}[\| g_k^{(sE+i)}\|_2^2]+\frac{2\eta^2}{KM}\sum_{i=1}^{r-1}\sum_{l=1}^M\mathbb{E}[\|\tilde{g}_l^{(sE+i)}-g_l^{(sE+i)}\|_2^2]+\frac{2(r-1)\eta^2}{M}\sum_{i=1}^{r-1}\sum_{l=1}^M\mathbb{E}[\| g_l^{(sE+i)}\|_2^2]\\
        &\overset{\text{\cfive}}\le 2(r-1)\eta^2\frac{G^2}{b}+2(r-1)\eta^2\sum_{i=1}^{r-1}\mathbb{E}[\| g_k^{(sE+i)}\|_2^2]+\frac{2(r-1)\eta^2}{K}\frac{G^2}{b}+\frac{2(r-1)\eta^2}{M}\sum_{i=1}^{r-1}\sum_{l=1}^M\mathbb{E}[\| g_l^{(sE+i)}\|_2^2]\\
        &=\frac{2(K+1)(r-1)\eta^2G^2}{Kb}+2(r-1)\eta^2\sum_{i=1}^{r-1}\mathbb{E}[\| g_k^{(sE+i)}\|_2^2]+\frac{2(r-1)\eta^2}{M}\sum_{i=1}^{r-1}\sum_{l=1}^M\mathbb{E}[\| g_l^{(sE+i)}\|_2^2],
    \end{flalign*}
    where \text{\cone} is due to the triangle inequality $\|a+b\|_2\le\|a\|_2+\|b\|_2$ and the Cauchy—Schwarz inequality, \text{\ctwo} follows from $\mathbb{E}[w^2]=\mathbb{E}[(w-\mathbb{E}[w])^2]+(\mathbb{E}[w])^2$, \text{\cthree} is due to the independent mini-batch sampling and unbiased gradient estimation, \text{\cfour} uses Cauchy—Schwarz inequality again, and \text{\cfive} is due to bounded local variance assumption of local stochastic gradients (Assmption~\ref{asp-localvar}). 
    
    Next, we let $(s,r,k)$ vary and bound $\sum\limits_{s=0}^{\lfloor\frac{T-1}{E}\rfloor}\sum\limits_{r=1}^{E}\sum\limits_{k=1}^M\mathbb{E}[\|\theta^{(sE+r-1)}-\theta_k^{(sE+r-1)}\|_2^2]$.
    \begin{flalign*}
        &\sum_{s=0}^{\lfloor\frac{T-1}{E}\rfloor}\sum_{r=1}^{E}\sum_{k=1}^M\mathbb{E}[\|\theta^{(sE+r-1)}-\theta_k^{(sE+r-1)}\|_2^2]\\
        &=\sum_{s=0}^{\lfloor\frac{T-1}{E}\rfloor}\sum_{r=1}^{E}\sum_{k=1}^M\Bigg(\frac{2(K+1)(r-1)\eta^2G^2}{Kb}+2(r-1)\eta^2\sum_{i=1}^{r-1}\mathbb{E}[\| g_k^{(sE+i)}\|_2^2]+\frac{2(r-1)\eta^2}{M}\sum_{i=1}^{r-1}\sum_{l=1}^M\mathbb{E}[\| g_l^{(sE+i)}\|_2^2]\Bigg)\\
        &\le\frac{TME(K+1)\eta^2G^2}{Kb}+2\eta^2\sum_{r=1}^{E}(r-1)\sum_{s=0}^{\lfloor\frac{T-1}{E}\rfloor}\sum_{k=1}^M\sum_{i=1}^{r-1}\mathbb{E}[\| g_k^{(sE+i)}\|_2^2]+\frac{2\eta^2}{M}\sum_{r=1}^{E}(r-1)\sum_{s=0}^{\lfloor\frac{T-1}{E}\rfloor}\sum_{k=1}^M\sum_{i=1}^{r-1}\sum_{l=1}^M\mathbb{E}[\| g_l^{(sE+i)}\|_2^2]\\
        &\le \frac{TME(K+1)\eta^2G^2}{Kb}+\eta^2E(E-1)\sum_{s=0}^{\lfloor\frac{T-1}{E}\rfloor}\sum_{i=1}^{E}\sum_{k=1}^M\mathbb{E}[\| g_k^{(sE+i)}\|_2^2]+\eta^2E(E-1)\sum_{s=0}^{\lfloor\frac{T-1}{E}\rfloor}\sum_{i=1}^{E}\sum_{l=1}^M\mathbb{E}[\| g_l^{(sE+i)}\|_2^2]\\
        &=\frac{TME(K+1)\eta^2G^2}{Kb}+2\eta^2E(E-1)\sum_{s=0}^{\lfloor\frac{T-1}{E}\rfloor}\sum_{i=1}^{E}\sum_{k=1}^M\mathbb{E}[\| g_k^{(sE+i)}\|_2^2]\\
        &=\frac{TME(K+1)\eta^2G^2}{Kb}+2\eta^2E(E-1)\sum_{t=1}^{T}\sum_{k=1}^M\mathbb{E}[\| g_k^{(t)}\|_2^2]
    \end{flalign*}
    Finally, we get
    \begin{flalign*}
         \frac{1}{TM}\sum_{t=1}^{T}\sum_{k=1}^M\mathbb{E}[\|\theta^{(t-1)}-\theta_k^{(t-1)}\|_2^2]
         &\le\frac{E(K+1)\eta^2G^2}{Kb}+\frac{2\eta^2E(E-1)}{TM}\sum_{t=1}^{T}\sum_{k=1}^M\mathbb{E}[\| g_k^{(t)}\|_2^2]\\
         &\le\frac{E(K+1)\eta^2G^2}{Kb}+\frac{2\eta^2\lambda E(E-1)}{T}\sum_{t=1}^{T}\mathbb{E}[\|\frac{1}{M}\sum_{k=1}^M g_k^{(t)}\|_2^2],
    \end{flalign*}
    where the last inequality is due to our bounded heterogeneity assumption (Assumption~\ref{asp-heterogeneity}).
\end{proof}

\subsection{Proof of Claim~\ref{cl-stgradnorm}}\label{sec:cl-stgradnorm}

\begin{proof}[Proof of Claim~\ref{cl-stgradnorm}]
    \begin{flalign*}
    &\quad\mathbb{E}[\|\tilde{g}^{(t)}\|_2^2]\\
    &=\mathbb{E}_{t-1}\mathbb{E}_{\mathcal{P}^{(t)}}\mathbb{E}_{\mathcal{B}^{(t)}}[\|\tilde{g}^{(t)}\|_2^2]\\
    &=\mathbb{E}_{t-1}\mathbb{E}_{\mathcal{P}^{(t)}}\mathbb{E}_{\mathcal{B}^{(t)}}[\|\tilde{g}^{(t)}-\mathbb{E}_{\mathcal{B}^{(t)}}[\tilde{g}^{(t)}]\|_2^2+\|\mathbb{E}_{\mathcal{B}^{(t)}}[\tilde{g}^{(t)}]\|_2^2]\\
    &=\mathbb{E}_{t-1}\mathbb{E}_{\mathcal{P}^{(t)}}\mathbb{E}_{\mathcal{B}^{(t)}}[\|(\frac{1}{K}\sum_{k\in\mathcal{P}^{(t)}}\tilde{g}_k^{(t)})-(\frac{1}{K}\sum_{k\in\mathcal{P}^{(t)}}g_k^{(t)})\|_2^2]+\mathbb{E}_{t-1}\mathbb{E}_{\mathcal{P}^{(t)}}[\|\frac{1}{K}\sum_{k\in\mathcal{P}^{(t)}}g_k^{(t)}\|_2^2]\\
    &\le\frac{1}{K^2}\mathbb{E}_{t-1}\mathbb{E}_{\mathcal{P}^{(t)}}\mathbb{E}_{\mathcal{B}^{(t)}}[\sum_{k\in\mathcal{P}^{(t)}}\|\tilde{g}_k^{(t)}-g_k^{(t)}\|_2^2]+\frac{1}{K}\mathbb{E}_{t-1}\mathbb{E}_{\mathcal{P}^{(t)}}[\sum_{k\in\mathcal{P}^{(t)}}\|g_k^{(t)}\|_2^2]\\
    &\le\frac{1}{K^2}\mathbb{E}_{t-1}[K\sum_{k=1}^M\frac{1}{M}\frac{G^2}{b}]+\frac{1}{K}\mathbb{E}_{t-1}[K\sum_{k=1}^M\frac{1}{M}\|g_k^{(t)}\|_2^2]\\
    &\le\frac{G^2}{Kb}+\lambda\mathbb{E}_{t-1}[\|\frac{1}{M}\sum_{k=1}^M g_k^{(t)}\|_2^2],
\end{flalign*}
where in the last inequality we use the bounded heterogeneity assumption (Assumption~\ref{asp-heterogeneity}).
\end{proof}

\subsection{Proof of Theorem~\ref{th-convergence}}\label{sec:th-convergence}
\begin{proof}[Proof of Theorem~\ref{th-convergence}]
    The choice of $\eta=\frac{1}{L\sqrt{\Gamma}T}$ comes by balancing the first and the last term in \eqref{eq-loss-bound}. With this choice of $\eta$, condition~\eqref{eq-rate_condition} becomes $\frac{E(E-1)}{T^2}<\frac{\Gamma}{2\lambda}-\frac{\sqrt{\Gamma}}{2T}$. By letting $T>\frac{2\lambda}{\sqrt{\Gamma}}$, it suffices to have $\frac{E}{T}<\frac{1}{2}\sqrt{\frac{\Gamma}{\lambda}}$ hold. Next we bound the mean-squared loss gradient norm. By the choice of $\eta$, we can calculate the sum of the first and the last term in \eqref{eq-loss-bound} as
    \begin{flalign*}
        \frac{2(F(\theta^{(0)})-F^*)}{\eta T}+\frac{\eta LG^2 T}{\rho_SMN}=\frac{3G\sqrt{L(F(\theta^{(0)})-F^*)}}{\sqrt{\rho_S MN}}.
    \end{flalign*}
    By taking the choice of $\eta$, we can calculate the the second term in \eqref{eq-loss-bound} as
    \begin{flalign*}
        \frac{\eta^2 L^2G^2E(\rho_CEM+T)}{\rho_SMN}
        &=L(F(\theta^{(0)})-F^*)\frac{E}{T}(\frac{\rho_C ME}{T}+1).
    \end{flalign*}
    Combining these results, we obtain the desired bound.
\end{proof}

\subsection{Proof of Corollary~\ref{coro-convergence}}\label{sec:coro-convergence}
\begin{proof}[Proof of Corollary~\ref{coro-convergence}]
    We start with the case of $E=T^{1-\alpha}$. Firstly, in order to have condition~\eqref{eq-rate_condition} hold, we require $T\ge\frac{2\lambda}{\sqrt{\Gamma}}$ and $\frac{E}{T}<\frac{1}{2}\sqrt{\frac{\Gamma}{\lambda}}$, which gives $T\ge\max\left\{\frac{2\lambda}{\sqrt{\Gamma}}, \left(2\sqrt{\frac{\lambda}{\Gamma}}\right)^{\frac{1}{\alpha}}\right\}$. Taking $E=T^{1-\alpha}$ into \eqref{eq-general-convergence}, we obtain
    \begin{equation}
        \frac{1}{T}\sum_{t=1}^{T}\mathbb{E}[\|\nabla F(\theta^{(t-1)})\|_2^2]\le \frac{3\sqrt{LG^2(F(\theta^{(0)})-F^*)}}{\sqrt{\rho_S MN}}+\frac{L(F(\theta^{(0)})-F^*)}{T^\alpha}(\frac{\rho_C M}{T^\alpha}+1).
    \end{equation}
    Furthermore, when $T$ is large enough such that $T>(\rho_C M)^\frac{1}{\alpha}$, we have $\frac{\rho_C M}{T^\alpha}<1$, which leads to the following
    \begin{flalign}
        \frac{1}{T}\sum_{t=1}^{T}\mathbb{E}[\|\nabla F(\theta^{(t-1)})\|_2^2]
        \le \frac{3G\sqrt{L(F(\theta^{(0)})-F^*)}}{\sqrt{\rho_S MN}}+\frac{2L(F(\theta^{(0)})-F^*)}{T^\alpha}.
    \end{flalign}
    Note that we have required $T\ge\left(2\sqrt{\frac{\lambda}{\Gamma}}\right)^\frac{1}{\alpha}$, we have
    \begin{flalign}
        \frac{2L(F(\theta^{(0)})-F^*)}{T^\alpha}
        &\le\frac{L(F(\theta^{(0)})-F^*)}{\sqrt{\lambda}}\sqrt{\frac{G^2}{L(F(\theta^{(0)})-F^*)\rho_S MN}}\\
        &=\frac{G\sqrt{L(F(\theta^{(0)})-F^*)}}{\sqrt{\lambda\rho_S MN}}\\
        &\le\frac{G\sqrt{L(F(\theta^{(0)})-F^*)}}{\sqrt{\rho_S MN}},
    \end{flalign}
    where the last inequality is due to the fact that $\lambda\ge 1$. Therefore, when $T\ge\max\left\{\frac{2\lambda}{\sqrt{\Gamma}}, \left(2\sqrt{\frac{\lambda}{\Gamma}}\right)^{\frac{1}{\alpha}}, \left(\rho_C M\right)^{\frac{1}{\alpha}}\right\}$, we have
    \begin{flalign}
        \frac{1}{T}\sum_{t=1}^{T}\mathbb{E}[\|\nabla F(\theta^{(t-1)})\|_2^2]\le \frac{4G\sqrt{L(F(\theta^{(0)})-F^*)}}{\sqrt{\rho_S MN}}=O\left(\frac{1}{\sqrt{\rho_S MN}}\right).
    \end{flalign}
    Next, we move onto the case of $E=c\cdot T$ for some constant $c<\frac{1}{2}\sqrt{\frac{\Gamma}{\lambda}}$ and $T\ge\frac{2\lambda}{\sqrt{\Gamma}}$. By direct calculation, we have
    \begin{flalign}
        \frac{1}{T}\sum_{t=1}^{T}\mathbb{E}[\|\nabla F(\theta^{(t-1)})\|_2^2]
        &\le \frac{3\sqrt{LG^2(F(\theta^{(0)})-F^*)}}{\sqrt{\rho_S MN}}+L(F(\theta^{(0)})-F^*)\frac{E}{T}(\frac{\rho_C ME}{T}+1)\\
        &\le \frac{3\sqrt{LG^2(F(\theta^{(0)})-F^*)}}{\sqrt{\rho_S MN}}+\frac{L(F(\theta^{(0)})-F^*)\rho_C M\Gamma}{4\lambda}+\frac{L(F(\theta^{(0)})-F^*)}{2}\sqrt{\frac{\Gamma}{\lambda}}\\
        &=\frac{3\sqrt{LG^2(F(\theta^{(0)})-F^*)}}{\sqrt{\rho_S MN}}+\frac{\rho_C G^2}{4\lambda\rho_S N}+\frac{\sqrt{LG^2(F(\theta^{(0)})-F^*)}}{2\sqrt{\lambda\rho_S MN}}\\
        &\le\frac{7\sqrt{LG^2(F(\theta^{(0)})-F^*)}}{2\sqrt{\rho_S MN}}+\frac{\rho_C G^2}{4\lambda\rho_S N},
    \end{flalign}
    where the last inequality follows from the fact of $\lambda\ge 1$. When $M<\frac{4\lambda^2L(F(\theta^{(0)})-F^*)\rho_S}{G^2\rho_C^2}N$, we obtain that $\frac{\rho_C G^2}{4\lambda\rho_S N}<\frac{\sqrt{LG^2(F(\theta^{(0)})-F^*)}}{2\sqrt{\rho_S MN}}$, which implies that
    \begin{flalign}
        \frac{1}{T}\sum_{t=1}^{T}\mathbb{E}[\|\nabla F(\theta^{(t-1)})\|_2^2]\le\frac{4\sqrt{LG^2(F(\theta^{(0)})-F^*)}}{\sqrt{\rho_S MN}}=O\left(\frac{1}{\sqrt{\rho_S MN}}\right).
    \end{flalign}
\end{proof}
\end{document}